\documentclass[letterpaper]{article} 
\usepackage{aaai24}  
\usepackage{times}  
\usepackage{helvet}  
\usepackage{courier}  
\usepackage[hyphens]{url}  
\usepackage{graphicx} 
\usepackage{natbib}  
\usepackage{caption} 
\usepackage{algorithm}
\usepackage{algorithmic}

\usepackage{amsmath}
\usepackage{amssymb}
\usepackage{mathtools}
\usepackage{amsthm}

\usepackage{subfig}
\usepackage{float}
\usepackage{multirow}
\usepackage{booktabs}

\usepackage{newfloat}
\usepackage{listings}
\DeclareCaptionStyle{ruled}{labelfont=normalfont,labelsep=colon,strut=off} 
\floatstyle{ruled}
\newfloat{listing}{tb}{lst}{}
\floatname{listing}{Listing}
\title{Component Fourier Neural Operator for Singularly Perturbed Differential Equations}
\author{
    Ye Li\textsuperscript{\rm 1},
    Ting Du\textsuperscript{\rm 2},
    Yiwen Pang\textsuperscript{\rm 1},
    Zhongyi Huang\textsuperscript{\rm 2}
}
\affiliations{
    \textsuperscript{\rm 1}Nanjing University of Aeronautics and Astronautics\\
    \textsuperscript{\rm 2}Tsinghua University\\
    yeli20@nuaa.edu.cn, dt20@mails.tsinghua.edu.cn, yiwenpang@nuaa.edu.cn, zhongyih@tsinghua.edu.cn

}

\usepackage{bibentry}

\begin{document}

\maketitle

\begin{abstract}
Solving Singularly Perturbed Differential Equations (SPDEs) poses computational challenges arising from the rapid transitions in their solutions within thin regions. The effectiveness of deep learning in addressing differential equations motivates us to employ these methods for solving SPDEs. In this manuscript, we introduce Component Fourier Neural Operator (ComFNO), an innovative operator learning method that builds upon Fourier Neural Operator (FNO), while simultaneously incorporating valuable prior knowledge obtained from asymptotic analysis. Our approach is not limited to FNO and can be applied to other neural network frameworks, such as Deep Operator Network (DeepONet), leading to potential similar SPDEs solvers. Experimental results across diverse classes of SPDEs demonstrate that ComFNO significantly improves accuracy compared to vanilla FNO. Furthermore, ComFNO exhibits natural adaptability to diverse data distributions and performs well in few-shot scenarios, showcasing its excellent generalization ability in practical situations.
\end{abstract}

\section{Introduction}
Singularly perturbed differential equations (SPDEs) serve as fundamental mathematical models in diverse physical phenomena, including fluid flows and material sciences \cite{roos2008robust}. Formally, SPDEs can be regarded as a distinct class of differential equations with a small positive parameter $\varepsilon$ appearing before the highest order derivative. These equations yield solutions that undergo rapid changes in thin regions, commonly referred to as boundary layers or inner layers, depending on their respective locations. Consequently, solving SPDEs poses significant challenges, both analytically and numerically.

With the surge of deep learning, efforts have been directed toward employing artificial neural networks for solving partial differential equations (PDEs) \cite{roos2008robust}, particularly in the field of physics-informed machine learning \cite{bar2019learning,greenfeld2019learning,karniadakis2021physics}. Notably, operator learning techniques like FNO \cite{li2020fourier} and DeepONet \cite{lu2021learning} have gained attention for their ability to learn operators between infinite-dimensional functional spaces. However, when addressing SPDEs, standard methods like vanilla FNO/DeepONet, and other neural network-based PDE solvers, face challenges. The occurrence of thin layers within SPDE solutions introduces stiffness-related features in neural network models.

In this study, we present the Component Fourier Neural Operator (ComFNO) for SPDEs. ComFNO extends FNO, tailored to boundary or inner layer phenomena. Unlike the explicit inclusion of physical equations, our method integrates prior knowledge from asymptotic analysis directly into the neural network architecture. ComFNO inherently differentiates between smooth and layer parts of SPDE solutions, enabling separate implicit learning. Our demonstrations reveal enhanced predictive accuracy, even in few-shot cases, and adaptability across diverse data distributions. ComFNO's versatility positions it as a promising tool. Our contributions are threefold:
\begin{itemize}
    \item We propose ComFNO, integrating asymptotic analysis insights into vanilla FNO's architecture.
    \item Our approach is not limited to FNO, offering a flexible framework applicable to SPDEs and compatible with alternative neural network architectures, like DeepONet.
    \item Experimental results across various SPDEs, encompassing one-dimensional, two-dimensional, and time-dependent equations, demonstrate significant reductions in mean, infinity norm, and residual error variance. This underscores the improved accuracy of ComFNO in addressing SPDEs, while its robustness is confirmed through empirical investigations involving multiple data distributions and few-shot cases.
\end{itemize}
\section{Related Work}
\subsection{Physics-Informed Machine Learning}
While pure data-driven machine learning has achieved breakthroughs in many fields, researchers have attempted to combine physical knowledge with machine learning to improve performance.
Physics-informed machine learning integrating seamlessly data and mathematical physics models (e.g., PDEs), has been one of the hotspots in current researches \cite{karniadakis2021physics,markidis2021old}.
This development, stemming from physics-informed neural networks (PINNs) and other neural network-based PDE solvers \cite{raissi2019physics,e2018deep,bar2019learning,rackauckas2020universal,yadav2021spde}, signifies a noteworthy evolution. 

PINNs employ tailored objective functions for diverse differential equations, utilizing limited datasets and even unsupervised learning. They harness inherent physical insights but may lack thorough data distribution analysis. However, PINN training presents unique challenges compared to conventional supervised learning.\cite{krishnapriyan2021characterizing,wang2022and}.

\subsection{Neural Operators}
Research into operator learning has recently surged, aiming to approximate mappings between infinite-dimensional functional spaces \cite{goswami2022physics}. This innovation eliminates the need for repetitive equation-solving with varying parameters, such as coefficients or source terms, promising significant speed-up compared to vanilla PINNs and traditional solvers. Contributions from the introduction of Deep Operator Network (DeepONet) \cite{lu2021learning} and Fourier Neural Operator (FNO) \cite{li2020fourier} are noteworthy. DeepONet harnesses the universal approximation theorem for operators \cite{chen1995universal}, accompanied by rigorous convergence analyses \cite{deng2022approximation,lanthaler2022error,de2022generic} and diverse applications \cite{lu2022comprehensive,wang2021learning,jin2022mionet,haghighat2021nonlocal}. However, it often demands an extensive dataset to enhance predictive performance. FNO boasts theoretical error estimates \cite{kovachki2021universal} and the capability to simulate various physical phenomena, including fluid flows \cite{li2021physics,rosofsky2022applications}, seismic waves \cite{yang2021seismic}, and material modeling \cite{you2022learning}. Although DeepONet and FNO perform comparably across several crucial PDEs \cite{lu2022comprehensive}, FNO is chosen in this study due to its superior cost-accuracy trade-off \cite{de2022cost}.
\subsection{Numerical and Neural Networks for Solving SPDEs}
Addressing SPDEs is a key concern in applied mathematics. Classical numerical techniques, including finite difference and finite element methods, have been extensively studied for such purposes \cite{roos2008robust,roos2022robust}. Yet, these methods require predefined grids, with accuracy tied to grid density. The rise of deep learning has led to the use of neural networks in tackling SPDEs \cite{tawfiq2014design}. Improvements in network architectures and training strategies have enhanced SPDE-solving capabilities \cite{liu2020legendre,greenfeld2019learning,simos2022neural,beguinet2022deep}. Nevertheless, neural network solvers for SPDEs reveal limitations in physics-informed machine learning methods like PINNs. Furthermore, these methods focus on equation solutions rather than solution operators. Developing operator solvers for SPDEs is particularly challenging due to the rapid transitions in solutions within thin regions.
\section{Problem Settings and Preliminaries}
\subsection{Singularly Perturbed Differential Equations}

Singularly perturbed differential equations (SPDEs) involve a small positive parameter, usually denoted as $\varepsilon$, which typically appears ahead of the highest-order derivative term. As $\varepsilon$ approaches 0, the solution's derivative (or higher-order derivatives) can tend towards infinity within specific regions. Meanwhile, the solution (or its derivative) undergoes significant changes in these regions, while displaying regular behavior away from them. These regions are termed boundary layers or inner layers, depending on their relative positions.

\begin{figure}[H]
    \centering
    \includegraphics[width=0.9\columnwidth]{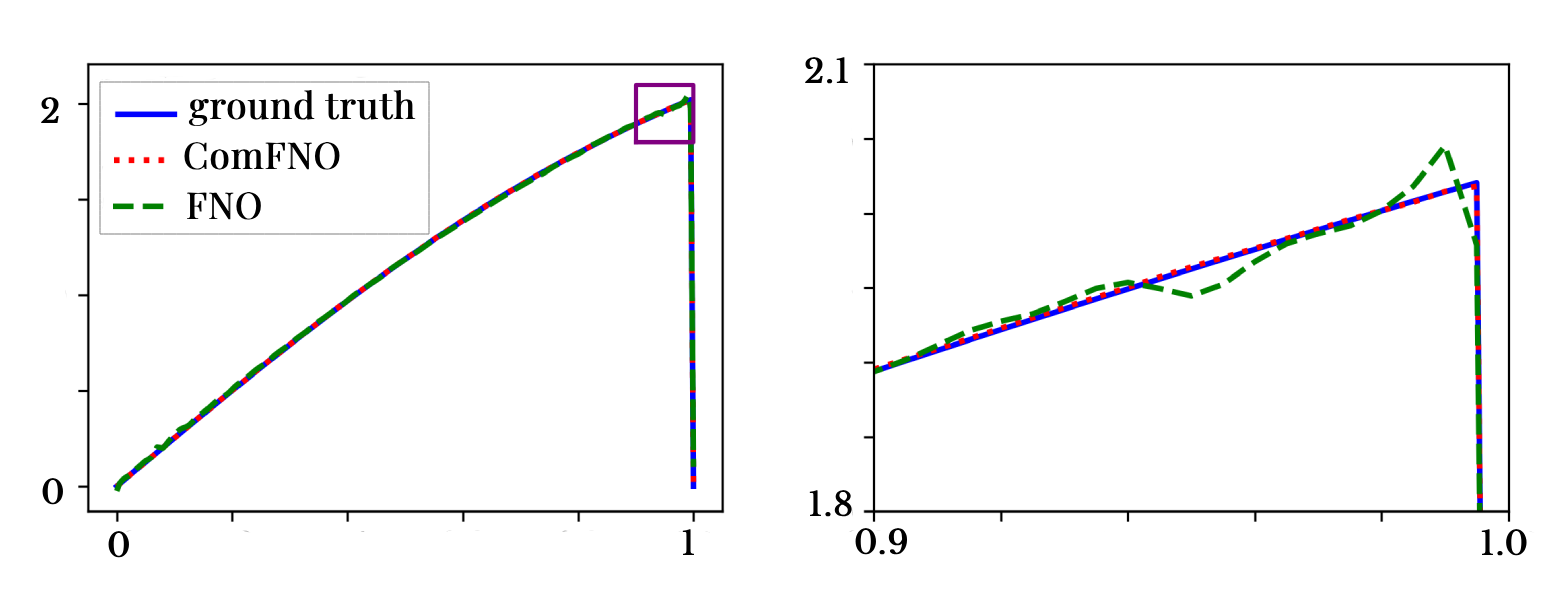}
    \caption{(left) Ground truth and predictions for FNO and ComFNO. Here we have $b(x)=x+1$, $c(x)=0$ and $\varepsilon=0.001$ in Eq. \eqref{ode_example}. We take 900 distinct $f(x)$ for training and a random $f(x)$ for testing. (right) Zoomed in curves near the boundary layer. We can see that, the true solution of Eq. \eqref{ode_example} has a boundary layer at $x=1$. }
    \label{fig:ordinary_1_predicts}
\end{figure}

We illustrate the concept using the one-dimensional convection-diffusion equation as an example—a straightforward yet highly significant category of SPDEs—presented in the following form:
\begin{equation}\label{ode_example}
\left\{
\begin{aligned}
    &-\varepsilon u''+b(x)u'+c(x)u=f(x),\quad x\in\left(0,1\right),\\
    &u(0)=u(1)=0,
\end{aligned}\right.
\end{equation}
where $\varepsilon$ is a very small positive parameter. The higher-order term $u''$ delineates diffusion, while the first-order term $u'$ signifies convection. Owing to the existence of this small parameter $\varepsilon$, the solution to such an equation tends to exhibit singular behavior, as demonstrated in Figure \ref{fig:ordinary_1_predicts}.

\subsection{Fourier Neural Operator}

Fourier neural operator (FNO) is derived from the kernel neural operator, which replaces the operator with the Fourier operator \cite{li2020fourier}. The Fourier layer, constituting a fundamental building block of FNO, drives the updates $v_t\mapsto v_{t+1}$ as follows:
\begin{equation*}
    v_{t+1}(x):=\sigma\left(Wv_t(x)+\left(\mathcal{K}(a;\theta)v_t\right)(x)\right).
\end{equation*}
Here, $W$ represents a linear transformation, and $\sigma$ denotes a nonlinear activation function. The parameterized neural network $\mathcal{K}(a;\theta)$ is characterized by the subsequent formulation:
\begin{equation*}
    \left(\mathcal{K}(\phi)v_t \right)(x)=\mathcal{F}^{-1}\left(R_\theta\cdot(\mathcal{F}v_t)\right)(x).
\end{equation*}
In this context, $\mathcal{F}$ stands for the Fourier transform, while $\mathcal{F}^{-1}$ signifies its inverse. These transformations are detailed as:
\begin{equation*}
    \begin{aligned}
\left(\mathcal{F}f\right)\left(k\right)&=\int_Df(x)e^{-2i\pi\left<x,k\right>}dx,\\
\left(\mathcal{F}^{-1}f\right)\left(x\right)&=\int_Df(k)e^{2i\pi\left<x,k\right>}dx.
    \end{aligned}
\end{equation*}
\section{Method}
\subsection{Asymptotic Analysis}
Our framework builds upon the principles of singular perturbation theory and asymptotic expansion methods. The function $u_{as}$ represents an asymptotic expansion of order $m$ of $u$ if there exists a constant $C$ such that for all $x\in [0,1]$ and sufficiently small $\varepsilon$, the following inequality holds: 
\[|u(x)-u_{as}(x)|\le C\varepsilon^{m+1}.\]

In numerous classes of SPDEs, their solutions or asymptotic expansions often exhibit a decomposition into two distinct components. One part characterizes the solution's behavior within boundary or inner layers referred to as the ``layer part"—while the other part pertains to the solution's behavior outside these regions, termed the ``smooth part." Within this manuscript, our focus primarily centers on exponential-type layers, which occur frequently and exert substantial influence on the solution. For these exponential-type layers, the layer part shares a significant relationship with exponential functions, thereby furnishing pivotal insights that underpin our innovative framework.

Extensive research has focused on the asymptotic analysis of SPDEs \cite{roos2008robust,becher2015richardson}. We will now present some notable contributions in this domain, with illustrative examples further elucidating boundary and inner layer phenomena in SPDEs available in the Appendix.
\subsubsection{Ordinary Differential Equations}

To begin, we assume that all the forthcoming equations involve sufficiently smooth coefficient functions and source terms. The term "reduced solutions" refers to solutions of the reduced problems obtained by setting $\varepsilon=0$ in the SPDEs.

Let's commence with a simple convection-diffusion equation:
\begin{equation}
\left\{\begin{array}{lr}
   -\varepsilon u''+b(x)u'+c(x)u=f(x),   & x\in(0,1), \\
    u(0)=u(1)=0. & 
\end{array}\right.\label{equ:ode1}
\end{equation}
The solution of Eq. \eqref{equ:ode1} typically exhibits an exponential boundary layer at $x=1$ when $b(x)>0$ on $[0,1]$, and a similar layer at $x=0$ when $b(x)<0$. In cases where $b(x)$ has zeros on $[0,1]$, we refer to it as the turning point problem, which will be addressed separately later. The conditions $b<0$ and $b>0$ are mutually equivalent, as the change of variable $x\mapsto 1-x$ transforms the problem from one formulation to the other. Focusing on the case when $b(x)\ge \beta>0$, the solution $u$ has an asymptotic expansion of order $m$ in the following form:
\begin{equation}
u_{as}(x)=\sum_{\nu=0}^m\varepsilon^\nu u_\nu(x)+\sum_{\mu=0}^m\varepsilon^\mu v_\mu\left(\frac{1-x}{\varepsilon}\right).\label{equ:ode1_expansion}
\end{equation}

The functions $u_{\nu}\ (\nu=0,1,\dots,m)$ and $v_{\mu}\ (\mu=0,1,\dots,m)$ in Eq. \eqref{equ:ode1_expansion} are obtained through matched asymptotic expansion \cite{eckhaus2011matched}, a widely-used technique in asymptotic analysis. This method identifies $u_0$ as the reduced solution, and $v_0\left(\frac{1-x}{\varepsilon}\right)=-u_0(1)\exp\left(-b(1)\frac{1-x}{\varepsilon}\right)$, yielding an asymptotic expansion with the following estimate:
\begin{equation}
   |u(x)-(u_0+v_0)|\le C\varepsilon,\label{expansion_ode1}
\end{equation}
prompting the integration of exponential operations in FNO.

In Eq. \eqref{expansion_ode1}, $u_0$ represents the smooth part of the asymptotic expansion, capturing the function's smooth behavior across most regions except the boundary layer, while $v_0$ serves as the the layer part, acting as a correction within the boundary layer region. Beyond the asymptotic expansion, the solution $u$ to Eq. \eqref{equ:ode1} can be further decomposed into two parts: the smooth part denoted by $S$, which satisfies
\begin{equation*}
\lvert S(x)\rvert \le C,
\end{equation*}
and the layer part denoted by $E$, which satisfies
\begin{equation*}
\lvert E(x)\rvert \le C\varepsilon^{-l}\exp\left(-\frac{\beta(1-x)}{\varepsilon}\right).
\end{equation*}

In the context of turning point problems, isolated points where the coefficient of $u'$ vanishes are referred to as turning points. In this study, we focus on the scenario of a single turning point located within the interior of the domain, without loss of generality, where the differential equation is defined on the interval $(-1,1)$ with the turning point situated at $x=0$. Thus, we investigate the following equation by assuming $b(x)\ne 0$:
\begin{equation}
\left\{
\begin{aligned}
&-\varepsilon u''+xb(x)u'+c(x)u=f(x),\quad x\in\left(-1,1\right),\\
&u(-1)=u(1)=0.
\end{aligned}\right.\label{equ:ode_tp}
\end{equation}

It is crucial to emphasize that the solution $u(x)$ may demonstrate singular behavior at the turning point $x=0$ and the boundary points $x=-1$ and $x=1$ (for further details, refer to the Appendix). For our analysis, we consider the case where $b(x)\ge\beta>0$ on $[-1, 1]$, resulting in the emergence of two distinct boundary layers at $x=-1$ and $x=1$. For this particular case, the solution's asymptotic expansion is given by
\begin{equation*}
    u_{as}=u_0+v_0+w_0,\ \text{with}\ |u(x)-u_{as}(x)|\le C\varepsilon,
\end{equation*}
where $u_0$ is the reduced solution, $v_0$ and $w_0$ are defined as follows:
\begin{equation*}
\begin{aligned}
    v_0(x)=(u(1)-u_0(1))\exp\left(-b(1)\frac{1-x}{\varepsilon}\right),\\
    w_0(x)=(u(-1)-u_0(-1))\exp\left(b(-1)\frac{1+x}{\varepsilon}\right).
\end{aligned}
\end{equation*}
\subsubsection{Partial Differential Equations}
In the context of parabolic partial differential equations in the space-time domain $Q=(0,1)\times(0,T]$, the initial-boundary value problem is described by the following equation:
\begin{equation}
\left\{
    \begin{aligned}
    &u_t-\varepsilon u_{xx}+b(x,t)u_x+d(x,t)u=f(x,t),\ &(x,t)\in Q,\\
    &u(x,0)=s(x),\quad&0\le x\le 1,\\
    &u(0,t)=q_0(t),\quad&0< t\le T,\\
    &u(1,t)=q_1(t),\quad&0< t\le T.
    \end{aligned}\right.\label{pde:ib}
\end{equation}

In cases where $b > 0$, the solution $u$ displays smooth behavior across most of domain $Q$. However, near the boundary $x = 1$ of $Q$, the solution typically manifests a boundary layer. For fixed $t > 0$, this layer's behavior relative to $x$ resembles that in Eq. \eqref{equ:ode1}. In this context, the solution $u(x,t)$ can be decomposed as:
\begin{equation*}
    u(x,t)=u_0(x,t)-\tilde u_0(1,t)e^{-b(1,t)(1-x)/\varepsilon}+w(x,t).
\end{equation*}
Here, $u_0$ is the reduced solution, $|\tilde u_0|=\mathcal{O}(1)$, and $\lvert w(x,t)\rvert\le C\varepsilon^{1/2}$.

For a boundary value problem of an elliptic partial differential equation in the spatial domain $\Omega=(0,1)\times(0,1)$, given by
\begin{equation}
\left\{
    \begin{aligned}
    &-\varepsilon\Delta u+\textbf{b}(x,y)\cdot\nabla u+c(x,y)u =f(x,y),\ &\text{in}\ \Omega,\\
    &u(x,y)=0,\  &\text{on}\ \partial\Omega.
    \end{aligned}\right.\label{pde:2d}
\end{equation}
Under the assumption of $\textbf{b}=(b_1,b_2)>0$ (specifically, with $b_1>0$ and $b_2>0$), the emergence of two exponential boundary layers is evident, situated at both $x=1$ and $y=1$. The interplay of these two boundary layers at the coordinate $(1,1)$ necessitates the introduction of a corner layer correction. The asymptotic expansion of $u$ is formulated as:
\begin{equation}
    \begin{aligned}
    u_{as}(x,y)&:=u^*_{as}(x,y)+v^*_{as}(x,y),\\
    u_{as}^*(x,y)&:=u_0(x,y)-u_0(1,y)e^{-b_1(1,y)\frac{1-x}{\varepsilon} }\\
    &-u_0(x,1)e^{ -b_2(x,1)\frac{1-y}{\varepsilon} }.&\\
    v^*_{as}(x,y)&:=u_0(1,1)e^{-b_1(1,1)b_2(1,1)\frac{1-x}{\varepsilon}\frac{1-y}{\varepsilon} }.
    \end{aligned}\label{expansion:2d}
\end{equation}
Here, $u_0$ is the reduced solution and the following estimation holds: 
\begin{equation*}
    |u(x)-u_{as}(x)|\le C\varepsilon.
\end{equation*}

\subsection{Improved Network Structure}

In the preceding section, we introduced asymptotic expansions and solution decompositions for diverse classes of SPDEs. These solutions, including their asymptotic expansions, can be partitioned into two components: the smooth part and the layer part. Remarkably, their solutions exhibit exponential boundary layers,  with the layer parts linked to exponential functions. Specifically, it is evident that in cases where the spatial dimension is one-dimensional, encompassing ordinary differential equations (ODEs) such as Eq. \eqref{equ:ode1}, \eqref{equ:ode_tp}, and partial differential equations (PDEs) like Eq. \eqref{pde:ib}, the layer part demonstrates behavior akin to $\exp(-c(x_0-x)/\varepsilon)$ as $x$ approaches $x_0$. Here, $x_0$ signifies the location of a boundary or inner layer, and $c$ represents a constant for ODEs and a function for PDEs. For instance, in Eq. \eqref{equ:ode_tp}, the solution reveals two exponential boundary layers at $x=-1$ and $x=1$, with the layer parts exhibiting behavior resembling $\exp(c(1-x)/\varepsilon)$ near $x=1$ and $\exp(c(-1-x)/\varepsilon)$ near $x=-1$, respectively.

\begin{figure*}[htbp]
    \centering
    \includegraphics[width=0.75\textwidth]{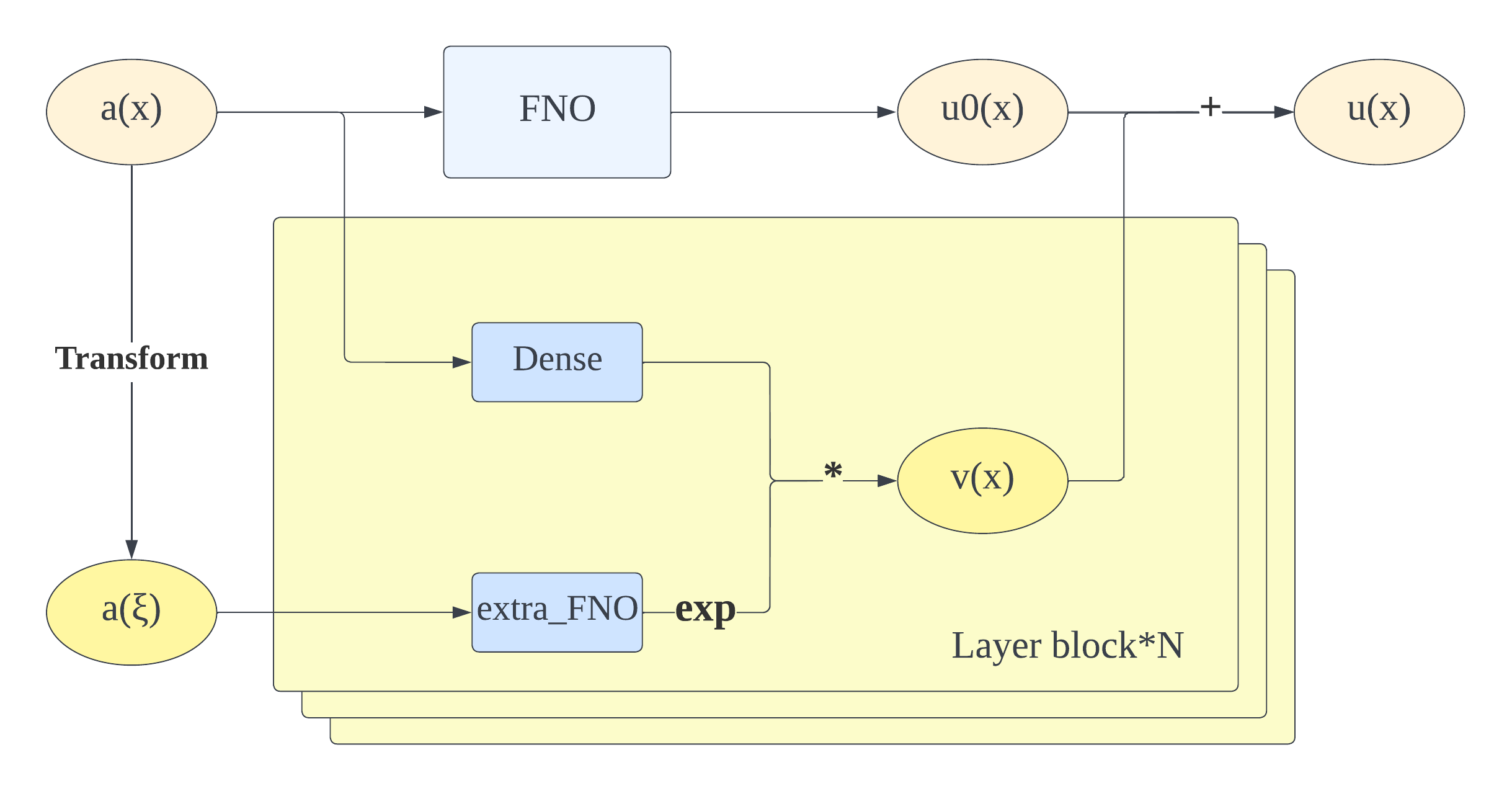}
    \caption{Architecture of ComFNO. $a(x)$ and $u(x)$ represent the input of model and solution of the problem, respectively. Both ``FNO" and ``extra\_FNO" represent Fourier Neural Operators, with the latter being smaller. An ``exp" function follows ``extra\_FNO," indicating an exponential operation applied to its output. ``Dense" corresponds to a shallow neural network that learns the coefficients of the exponential function. The layer block's input comprises both $a(x)$ and $a(\xi)$, the latter involving a coordinate transformation to accommodate scaling in layers. For instance, when encountering a boundary or inner layer at $x=x_0$, the use of $\xi=(x_0-x)/\varepsilon$ is advantageous.}
    \label{fig:spno_model}
\end{figure*}

This insight prompts the extension of vanilla FNO to the construction of the Component Fourier Neural Operator (ComFNO) illustrated in Fig. \ref{fig:spno_model}. By incorporating exponential operations and the coordinate transformation $x\mapsto\xi=(x_0-x)/\varepsilon$ to account for scaling in layers, ComFNO integrates prior knowledge of layer locations and types, thereby enhancing the existing operator learning framework. This approach involves two steps: (1) employing the ``FNO" block to capture the smooth parts, and (2) employing layer blocks to learn specific layer-related information. For ComFNO with $N$ layer blocks, the model can be succinctly represented by the equation:
\begin{equation}
\text{ComFNO}=\text{FNO}_0+\sum\limits_{i=1}^N\text{NN}_i*\exp(\text{FNO}_i).
\end{equation}
Here, $\text{FNO}_i$ (i=0,1,\dots,N) signifies the FNO model, while $\text{NN}_i$ (i=1,2,\dots,N) represents shallow neural networks. $\text{FNO}_0$ corresponds to the ``FNO" block in Fig. \ref{fig:spno_model}, while $\text{NN}_i*\exp(\text{FNO}_i)$ corresponds to the $N$ layer blocks in ComFNO, where $\text{NN}_i$ corresponds to the ``Dense" block and $\text{FNO}_i$ corresponds to the ``extra\_FNO" block.

For problems with spatial dimensions not less than 2, such as the two-dimensional PDE Eq. \eqref{pde:2d}, the solution exhibits two exponential boundary layers along $x=1$ and $y=1$, displaying similar behavior to $\exp({c(1-x)/\varepsilon})$ near $x=1$ and $\exp({c(1-y)/\varepsilon})$ near $y=1$. However, their overlapping at the coordinate $(1,1)$ gives rise to a corner layer. In theory, one could incorporate a layer block in ComFNO to rectify inaccuracies near the corner $(1,1)$. Yet, each addition of a layer block results in an increase in network complexity, necessitating a trade-off. Interestingly, we observe that the corner layer part $v_{as}^*$ in Eq. \eqref{expansion:2d} can also be expressed as $\exp({c(1-x)/\varepsilon})$ or $\exp({c(1-y)/\varepsilon})$ with $c$ as a function. Hence, the corner treatment is omitted, as the incorporation of two layer blocks is anticipated to rectify inaccuracies near the corner $(1,1)$. The subsequent section's experiment will demonstrate its effectiveness.

\section{Experiments}  

In this section, we apply ComFNO to a wide range of SPDEs, including both ordinary and partial differential equations, as well as scenarios involving multiple data distributions and few-shot cases. Furthermore, we conduct a comparative analysis of the experimental results with those obtained using FNO.

Subsequently, we detail the default experimental setup, where, unless explicitly stated, experiment parameters are established as follows. The parameter $\varepsilon$ in SPDEs remains set at $1 \times 10^{-3}$. Our aim is to learn the mapping $f\mapsto u$, where $f$ represents the source term. The training dataset consists of $900 \times 201$ tuples $(f,u)$, with 900 $f$ samples independently drawn from Gaussian random fields and used as inputs. Resolution on $[0,1]$ or $[-1,1]$ is fixed at 201. To derive $u$, we use high-precision numerical methods. For steady-state problems, which are independent of time, the upwind scheme on the Shishkin mesh is employed. For time-dependent problems, the Crank–Nicolson scheme on the Shishkin mesh is used \cite{roos2008robust}. More detailed configurations can be found in the Appendix.

After training the models, we will assess their performance on 100 different $f$ samples outside the training set, maintaining a resolution of 201. In the exposition of experimental findings, we will designate the high-precision numerical results as the ground truth, denoted by $u_g$, while the model predictions will be represented as $u_p$. Throughout this manuscript, we will visualize the prediction residuals of both FNO and ComFNO, namely, $u_p - u_g$. Detailed ground truth and prediction results will be presented in the Appendix.

\subsection{Ordinary Differential Equations}
\begin{figure}[htbp]
    \centering
    \includegraphics[width=0.8\columnwidth]{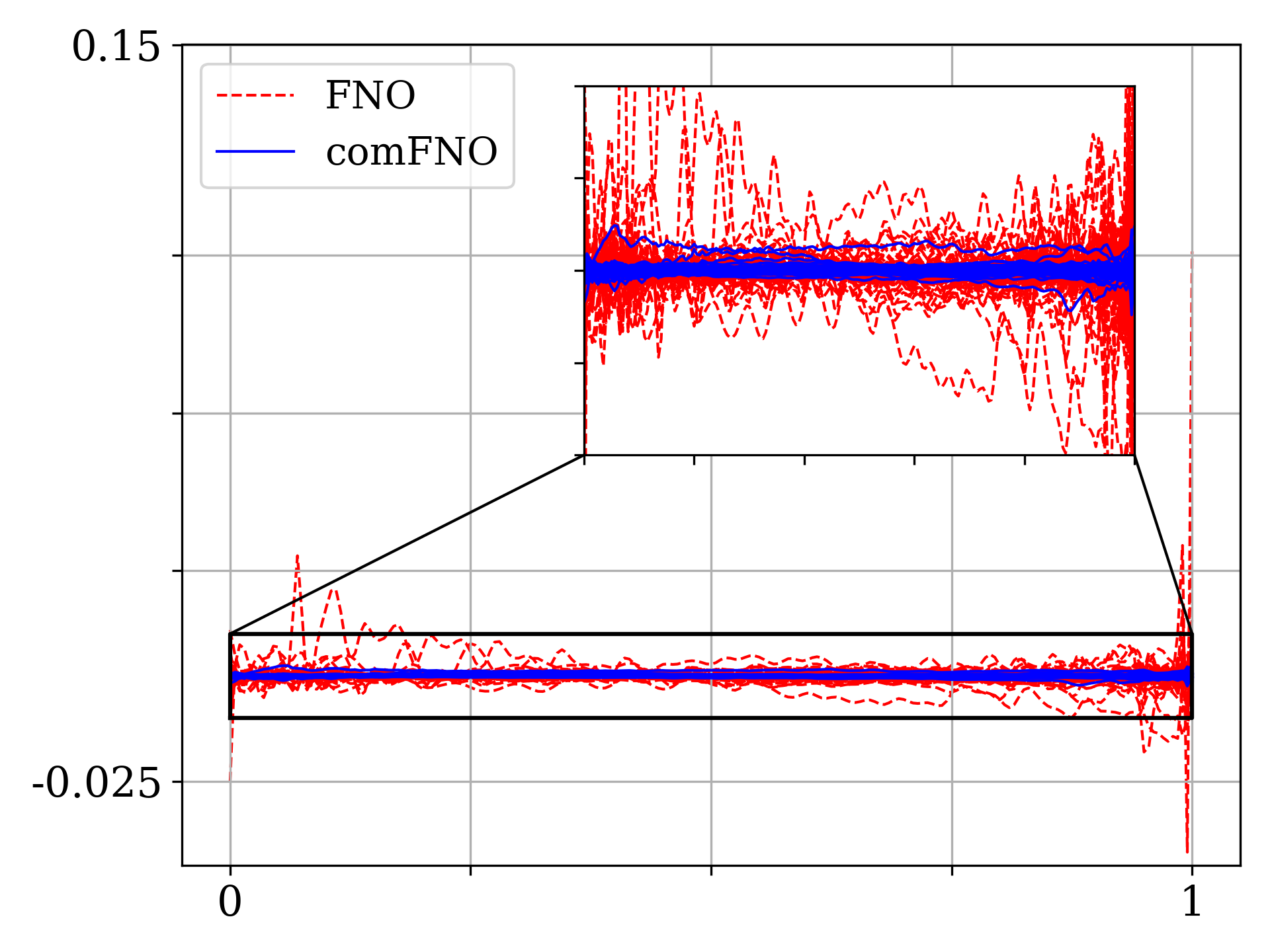}
    \caption{Performance of both FNO and ComFNO on Eq.\eqref{eq:ordinary_1} with $\varepsilon=0.001$. Both trained models are evaluated on 100 $f$ samples, and their resulting residual curves are depicted. (Subfigure): Zoomed-in view.}
    \label{fig:ordinary_2_residuals}
\end{figure}

For ordinary differential equations, we examine cases both with and without turning points. We begin with the following problem:
\begin{equation}
\left\{
    \begin{aligned}
    &-\varepsilon u''+(x+1)u'=f\quad x\in(0,1),\\
    &u(0)=u(1)=0.
    \end{aligned}\right.
    \label{eq:ordinary_1}
\end{equation}
\begin{figure}[htbp]
    \centering
    \includegraphics[width=0.8\columnwidth]{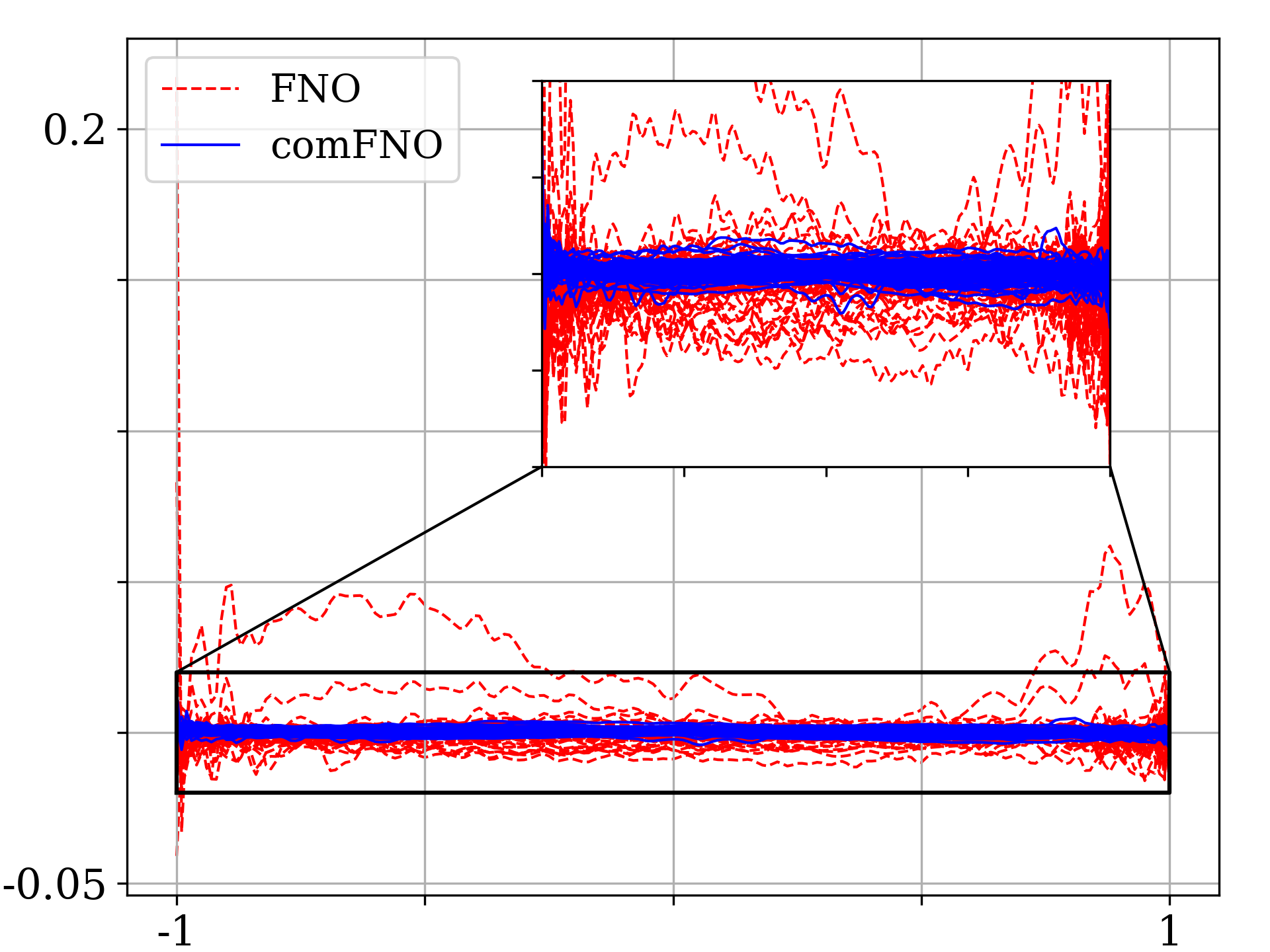}
    \caption{FNO and ComFNO performance on Eq.\eqref{eq:ordinary_2} with $\varepsilon=0.001$. Both trained models are evaluated on 100 $f$ samples, and their resulting residual curves are depicted. (Subfigure): Zoomed-in view.}
    \label{fig:ordinary_1_residuals}
\end{figure}
This problem's solutions feature an exponential boundary layer at $x=1$, leading us to employ a single layer block and incorporate a coordinate transformation $\xi=(1-x)/\varepsilon$ to incorporate $f(\xi)$ as one of the inputs to the layer block.

Li et al. \shortcite{li2020fourier} emphasizes FNO's resolution insensitivity, yielding improved results even at lower resolutions. Yet, as shown in Figure \ref{fig:ordinary_1_residuals}, FNO residuals are notably larger near $x=1$ compared to other regions. This stems from limited data near the boundary layer and significant solution variations as $\varepsilon$ approaches zero, potentially causing underfitting of the boundary layer.

Upon integrating a layer block, residuals decrease within the boundary layer. Surprisingly, this addition boosts accuracy near the boundary layer and reduces overall errors. Remarkably, for this case, our layer block structure harmonizes with the solution's inherent differential equation configuration.

Next we consider the case with turning point $x=0$:
\begin{equation}
\left\{
    \begin{aligned}
    &-\varepsilon u''+x(x+2)u'+u=f,\quad x\in(-1,1),\\
    &u(-1)=u(1)=0.
    \end{aligned}\right.
    \label{eq:ordinary_2}
\end{equation}
Given the exponential boundary layers at both $x=-1$ and $x=1$, two layer blocks in ComFNO are essential. These blocks receive inputs $(f(x),f((1-x)/\varepsilon))$ and $(f(x),f((-1-x)/\varepsilon))$, respectively. Fig. \ref{fig:ordinary_2_residuals} displays the prediction residuals of ComFNO and FNO, revealing smaller residuals for ComFNO across all regions. This highlights ComFNO's proficiency in effectively addressing turning point problems.

\subsection{Partial Differential Equations}
\begin{figure}[htbp]
    \centering
    \includegraphics[width=0.75\columnwidth]{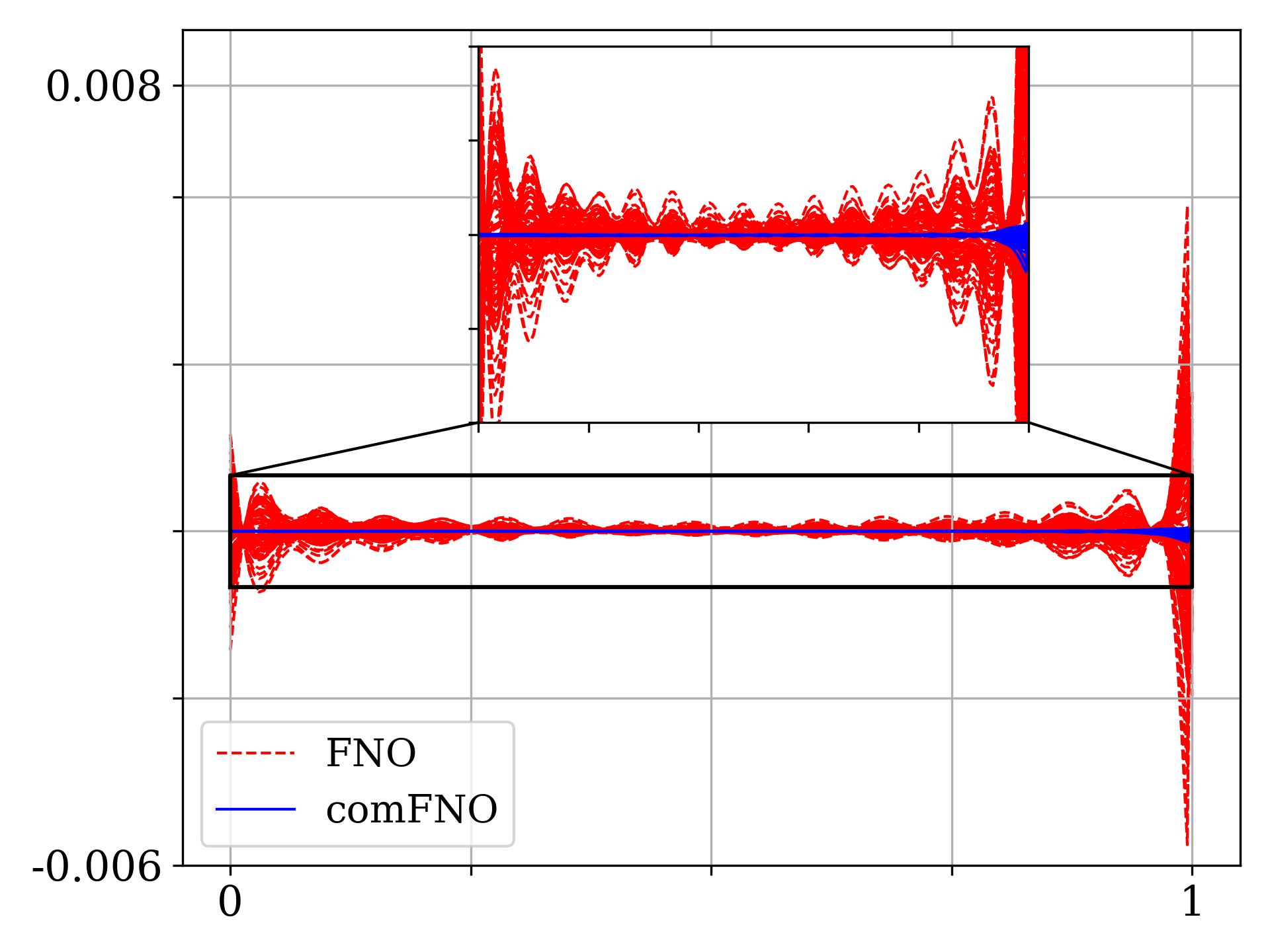}
    \caption{Performance of both FNO and ComFNO on Eq.\eqref{eq:partial_1} with $\varepsilon=0.001$. Both trained models are evaluated on 100 $f$ samples, and their resulting residual curves are depicted. (Subfigure): Zoomed-in view.}
    \label{fig:partial_1_residuals}
\end{figure}

Here we consider a parabolic differential equation in space-time domain:
\begin{equation}
\left\{\begin{array}{lr}
    u_t-\varepsilon u_{xx}+u_x+xu=0 &  (x,t)\in (0,1)\times(0,1]\\
    u(x,0)=f(x) & x\in [0,1]\\
    u(0,t)=u(1,t)=0,& t\in [0,1].
\end{array}\right.
\label{eq:partial_1}
\end{equation}
Our aim is to learn the mapping $f \mapsto u(\cdot,1)$. Incorporating a layer block with the input $(f(x),f((1-x)/\varepsilon))$ into ComFNO addresses the boundary layer near $x=1$. Fig. \ref{fig:partial_1_residuals} presents prediction residuals for both ComFNO and FNO.

\begin{figure}[htbp]
    \centering
    \subfloat[predicts on both two methods]{%
    \label{fig:partial_2_predicts}
    \includegraphics[width=0.9\columnwidth]{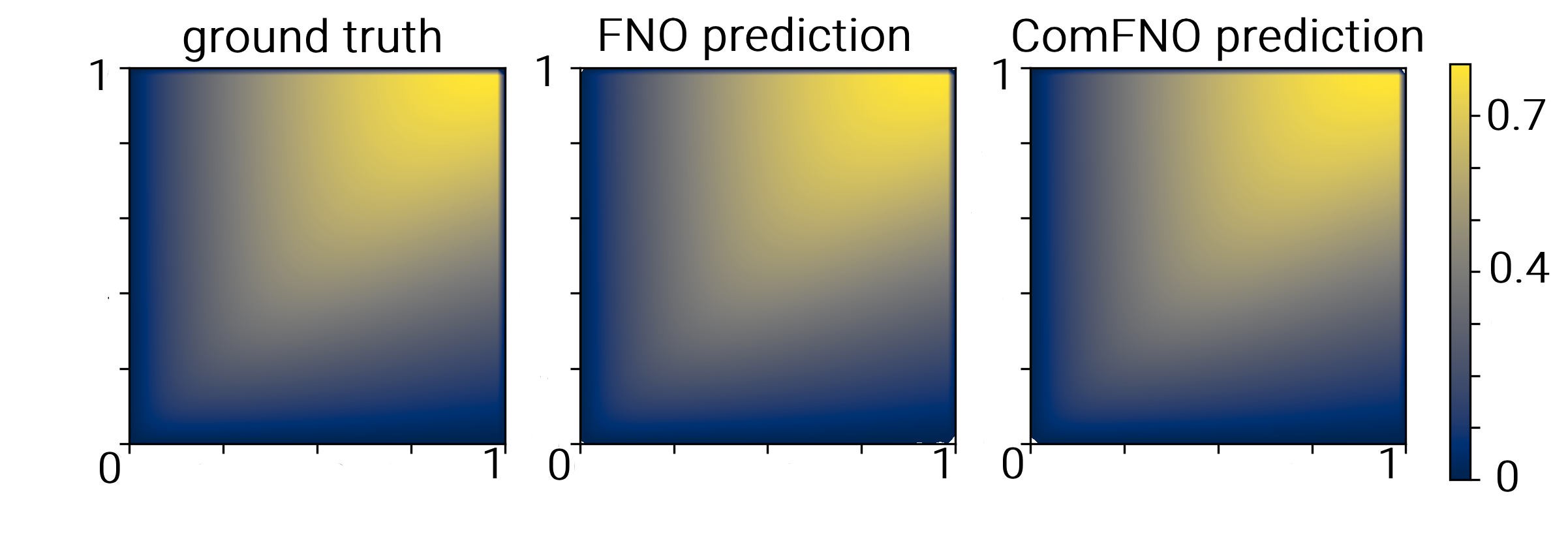}}\\
    \subfloat[absolute residuals on both two methods]{%
    \label{fig:partial_2_residuals}
    \includegraphics[width=0.8\columnwidth]{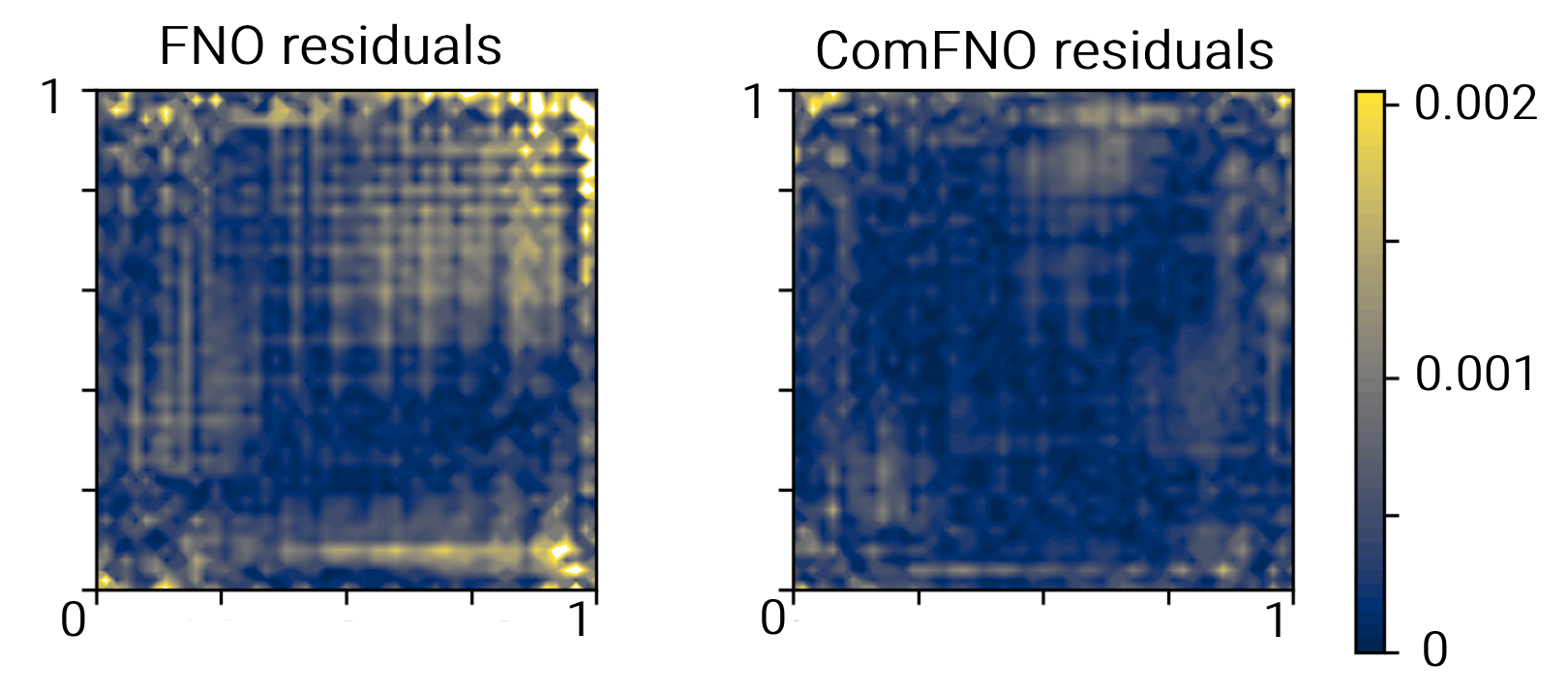}}
    \caption{Performance of FNO and ComFNO on Eq.\eqref{eq:partial_2} with $\varepsilon=0.001$. The training set consists of $900\times 51\times 51$ tuples $(f,u)$, including 900 independent $f$ samples, each with a resolution of $51\times 51$. Trained models are evaluated using a randomly selected $f$ sample. Absolute residuals for both models are presented, calculated as $|u_p-u_g|$, where $u_g$ is the ground truth and $u_p$ represents the model predictions.}
\end{figure}

Finally we consider an elliptic differential equation:
\begin{equation}
\left\{
\begin{array}{lr}
    -\varepsilon\Delta u+u_x+u_y+u=f(x), &  (x,y)\in(0,1)^2,\\
    u(0,y)=u(1,y)=0, & y\in [0,1]\\
    u(x,0)=u(x,1)=0, &x\in [0,1].\\
\end{array}\right.
\label{eq:partial_2}
\end{equation}
The boundary layer for this equation is present at $x=1$ and $y=1$. To address this, we utilize two layer blocks with inputs $(f(x,y),f((1-x)/\varepsilon),y)$ and $(f(x,y),f(x,(1-y)/\varepsilon))$ in ComFNO. In Fig. \ref{fig:partial_2_residuals}, our approach's efficacy throughout the entire region, not just the boundary, is evident.

\subsection{Multiple Distributions}
Vanilla FNO faces challenges when handling equations with various distinct $\varepsilon$ values concurrently, as differing $\varepsilon$ can result in diverse data distributions. To tackle this, an effective strategy is to include the parameter $\varepsilon$ as input or a prominent feature within the dataset, and the former method is used here. Our model follows a multi-input design, requiring the additional input $\varepsilon$ for effective coordinate transformation.

\begin{figure}[htbp]
    \centering
    \includegraphics[width=0.8\columnwidth]{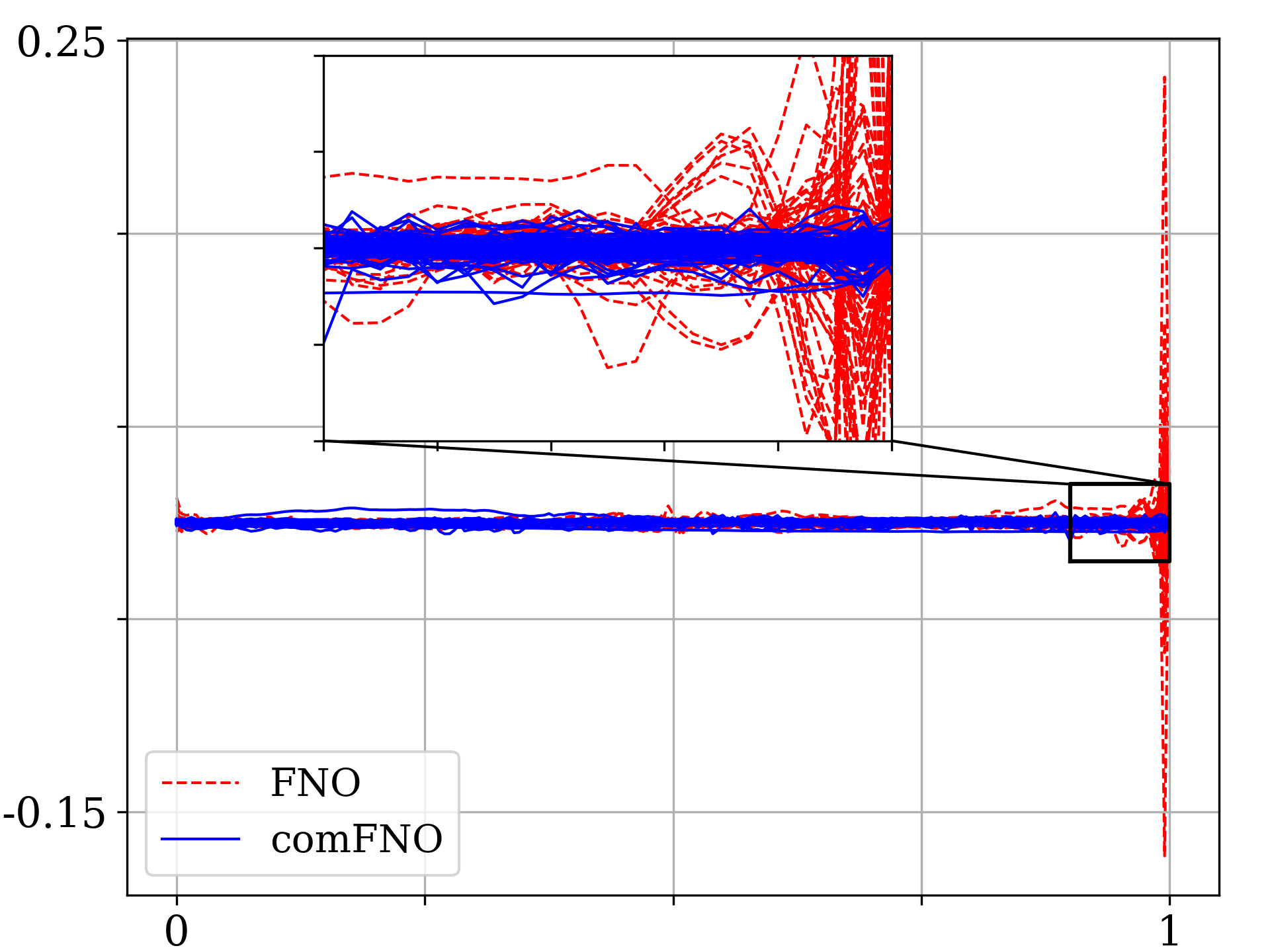}
    \caption{Performance of both FNO and ComFNO on Eq.\eqref{eq:ordinary_1}. Both models are evaluated using 100 $f$ samples with varying $\varepsilon$, and the corresponding residual curves are illustrated. (Subfigure): Zoomed-in view.}
    \label{fig:multiple_epsilon}
\end{figure}

\begin{table*}[pthb]
\centering
    \begin{tabular}{cccccccc} \toprule
        \multirow{2}{*}{Experiment} & \multicolumn{3}{c}{FNO} & &\multicolumn{3}{c}{ComFNO}\\ \cline{2-4}\cline{6-8}
         & mean &$\|\cdot\|_{L^{\infty}}$& var && mean&$\|\cdot\|_{L^{\infty}}$ & var\\ \hline
        1D(no turning point) & 5.0e-04& 3.8e-03 & 1.1e-06 && 1.0e-04 & 6.0e-04&  1.3e-08\\
        1D(turning point) & 1.6e-03& 8.5e-03 & 6.1e-06  && 6.0e-04& 2.1e-03& 1.9e-07 \\
        1D(initial-boundary) & 9.0e-05 & 1.7e-03& 4.8e-08  && 4.2e-06& 4.0e-05& 3.3e-11 \\
        2D & 5.0e-4& 1.67e-02 & 8.2e-07 && 2.0e-04 & 3.0e-03& 1.5e-07 \\
        multiple $\varepsilon$ & 7.0e-04& 1.7e-02 & 7.5e-06 && 6.0e-04& 2.1e-03 & 2.5e-07 \\
        few-shot & 1.2e-03& 4.5e-03 & 1.5e-06 && 3.0e-4 & 1.3e-03& 1.0e-07 \\
        \bottomrule
    \end{tabular}
    \caption{Mean, infinity norm, and variance of residuals on test set of all experiments.}
    \label{tab:tab_residuals}
\end{table*}

As an illustration, consider Eq. \eqref{eq:ordinary_1}. The training dataset is composed of $100 \times 100 \times 201$ triplets $(f,u)$. These triplets encompass 100 unique random functions $f$, each paired with 100 distinct $\varepsilon$ values ranging from 0.001 to 0.1 in increments of 0.001. The resolution is set at 201. Our main objective is to evaluate the ability of the two models to effectively fit this particular type of data.

We assess the model performance using 100 $f$ samples that beyond the training set. Each $f$ sample is associated with a distinct $\varepsilon$, ranging from 0.001 to 0.1. As shown in Fig. \ref{fig:multiple_epsilon}, the outcomes of FNO appear to lack certain data capturing layer structures. Conversely, ComFNO's results are more favorable, with minimal residuals observed in both boundary layers and other regions.

\subsection{Few-Shot Situation}
\begin{figure}[htbp]
    \centering
    \includegraphics[width=0.8\columnwidth]{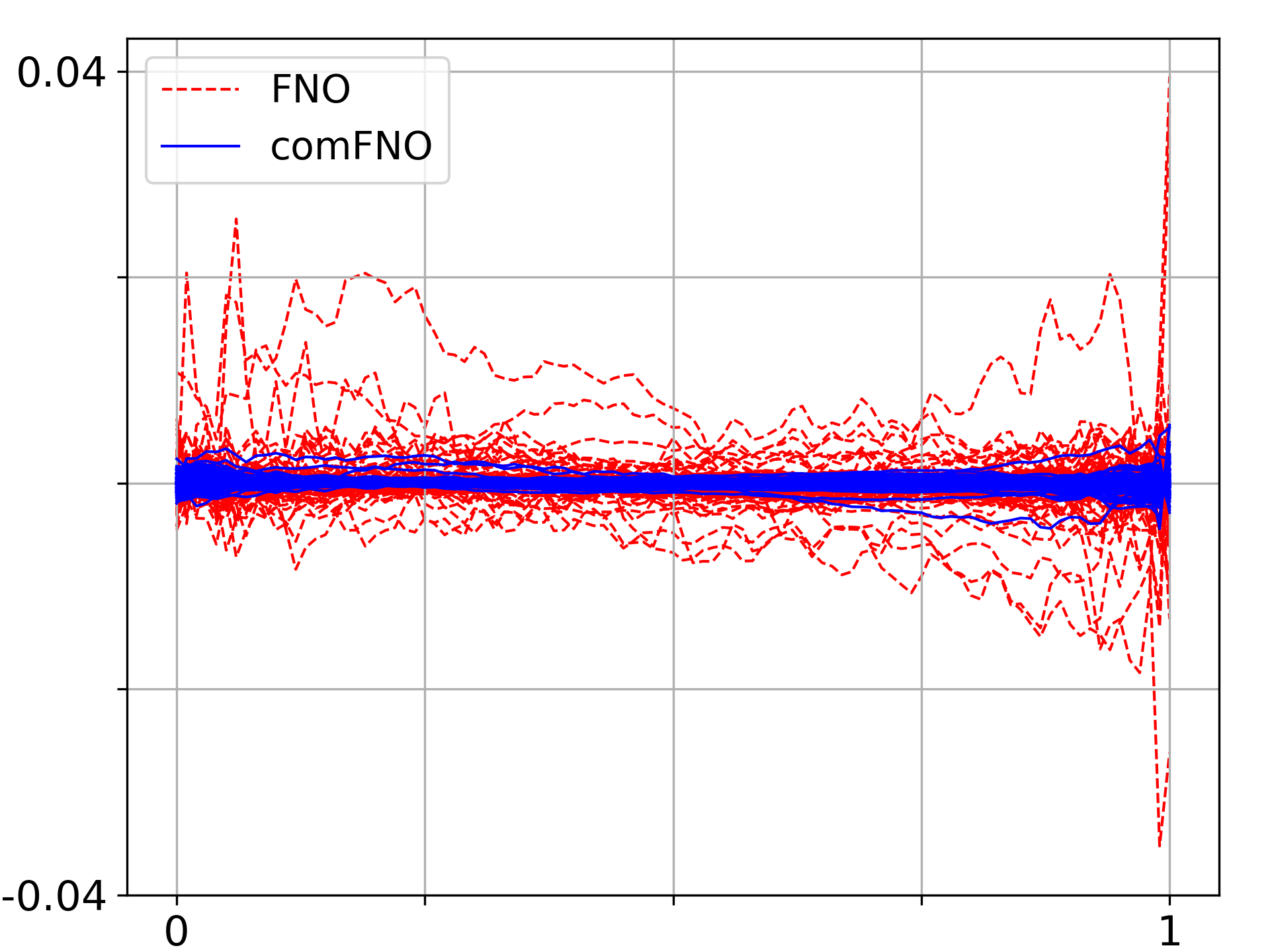}
    \caption{Residuals of FNO and ComFNO on Eq. \eqref{eq:ordinary_1} with $\varepsilon=0.001$. Trained on 100 samples and tested on 100 different samples.}
    \label{fig:fewshot_predicts}
\end{figure}

In this subsection, we primarily showcase the robustness of our method at varying sample sizes, highlighting the effectiveness of layer blocks in preserving sufficient physical information for successful model training, even with reduced data. We illustrate this using Eq. \eqref{eq:ordinary_1} as an example, where the training set is scaled down to $100\times101$ tuples $(f,u)$. The prediction residuals of both models are displayed in Fig.\ref{fig:fewshot_predicts}.

\subsection{Metrics}
In the previous experiments, we note that vanilla FNO's residuals display oscillations, mirroring the oscillatory nature of its predictions. Despite optimal training and minimal objective function values, these oscillations persist, compromising the model's reliability. For a more thorough comparative analysis of the two models, we introduce the metrics of mean, infinity norm, and variance for the residuals, as follows:
\begin{equation*}
    \begin{aligned}
    mean=\frac{1}{NM}\sum_{i=1}^{N}\sum_{j=1}^{M}\left|\hat{u}_{ij}-u_{ij}\right|,\\
    \|\cdot\|_{\infty}=\frac{1}{N}\sum\limits_{i=1}^N\mathop {\max }\limits_{0\ge j\ge M} \left|\hat{u}_{ij}-u_{ij}\right|,\\
    var=\frac{1}{NM}\sum_{i=1}^{N}\sum_{j=1}^{M}\left(\left|\hat{u}_{ij}-u_{ij}\right|-\frac{1}{M}\sum_{j=1}^{M}\left|\hat{u}_{ij}-u_{ij}\right|\right)^2,
    \end{aligned}
\end{equation*}
where $N$ is the number of samples, $M$ is the resolution. We computed mean, infinity norm, and variance of residuals for both FNO and ComFNO in all six experiments (Table \ref{tab:tab_residuals}), where smaller values denote superior model performance. Our innovative architecture demonstrates exceptional accuracy and effectively mitigates oscillations, as indicated by the indicators approaching zero.

\section{Conclusion}
This study introduces ComFNO, an innovative neural operator model tailored for addressing singularly perturbed differential equations (SPDEs). We initiate by presenting asymptotic analysis findings for various classes of SPDEs. Subsequently, we propose a unique layer block structure to enhance the training of conventional neural operators. Empirical results underscore the considerable improvement in prediction accuracy across boundary layer and other regions through the integration of layer blocks. Moreover, ComFNO exhibits superiority over vanilla FNO in specific instances, such as few-shot learning. Our model's versatility is demonstrated by testing it across multiple $\varepsilon$ values, highlighting its adaptability as a multi-input variant of FNO that can accommodate diverse distributions. Furthermore, our experiments validate ComFNO's capability to mitigate oscillations.

In our experiments, we specifically focused on the problem of exponential boundary layers. However, it is essential to underscore that our approach remains applicable to a broader range of scenarios. Once we possess prior knowledge about the location and type of layers, we can adapt the local variables $\xi$ accordingly and introduce additional operations within the layer blocks (e.g., incorporating exponential operations for exponential layers). This adaptability ensures the validity of our model for addressing inner layer problems or other types of boundary layer problems.

\section{Acknowledgments}

\bibliography{aaai24}
\newpage
\appendix
\onecolumn


\maketitle
\section{Experiment Settings}
In this section, we outline the experimental setup. For one-dimensional spatial problems, the training dataset includes $900 \times 201$ tuples $(f,u)$, while two-dimensional spatial scenarios encompass $900 \times 51\times51$ tuples $(f,u)$. In scenarios involving multiple data distributions, the training dataset consists of $100 \times 100 \times 201$ triplets $(f,\varepsilon,u)$, these triplets involve 100 distinct random functions $f$, each paired with 100 varying $\varepsilon$ values ranging from 0.001 to 0.1 in increments of 0.001. In the few-shot scenario, the training dataset comprises $100 \times 101$ tuples $(f,u)$. The input samples, designated as $f$, are drawn independently from Gaussian random fields, with the kernel function being a radial basis function kernel characterized by a fixed length-scale of 1. The corresponding outputs, represented by $u$, are obtained via high-precision numerical methodologies on a finer grid, utilizing interpolation at the predefined resolution.

In scenarios involving multiple data distributions, the test dataset consists of $100\times201$ triplets $(f,\varepsilon,u)$. Here, each $f$ is associated with a distinct $\varepsilon$, while maintaining a resolution of 201. In alternative cases, the test dataset comprises 100 distinct $f$ samples, distinct from those in the training dataset. The resolution remains consistent with the corresponding resolution of the training dataset.

In all conducted experiments, the loss functions employed uniformly adopt the $L_2$ relative error metric. The chosen optimizer for all minimization problems is Adam, accompanied by the consistent utilization of the GELU activation function. Further details concerning the remaining parameters pivotal for our result generation can be found in Table \ref{tab:comfno} and Table \ref{tab:fno}.

\begin{table}[pthb]
\centering
    \begin{tabular}{cccccc} \toprule
        Experiment/ComFNO & blockNum & LR & epoch& batch size\\ 
        \hline
        1D(no turning point) & 1& 0.001 & 500 &30\\
        1D(turning point) & 2& 0.001 & 500  &30 \\
        1D(initial-boundary) & 1 & 0.0001& 500  &30\\
        2D & 2& 0.001 & 1000 &20\\
        multiple $\varepsilon$ & 1& 0.001 & 1000 &50\\
        few-shot & 1& 0.001 & 2000 &25\\
        \bottomrule
    \end{tabular}
    \caption{Experimental parameters for ComFNO investigations. The term ``blockNum" denotes the quantity of layer blocks implemented within the architecture. ``LR" designates the learning rate employed, while ``epoch" signifies the count of training iterations performed.}
    \label{tab:comfno}
\end{table}

\begin{table}[pthb]
\centering
    \begin{tabular}{cccccc} \toprule
        Experiment/FNO &depth& LR & epoch& batch size\\ 
        \hline
        1D(no turning point) & 4& 0.001 & 500 &50\\
        1D(turning point) & 6& 0.001 & 500  & 50\\
        1D(initial-boundary) & 6 & 0.0001& 500  &50\\
        2D & 5& 0.001 & 1000 &50\\
        multiple $\varepsilon$ & 4& 0.001 & 1000 &50\\
        few-shot & 4& 0.002 & 2000 &50\\
        \bottomrule
    \end{tabular}
    \caption{Experimental parameters for FNO investigations. The term ``depth" denotes the quantity of Fourier layers implemented within the architecture. ``LR" designates the learning rate employed, while ``epoch" signifies the count of training iterations performed.}
    \label{tab:fno}
\end{table}

\section{Supplementary Experimental Results}
\subsection{One-dimensional Singularly Perturbed Differential Equations}
\begin{figure}[thbp]
	\centering
	\subfloat[]{\includegraphics[width=.4\columnwidth]{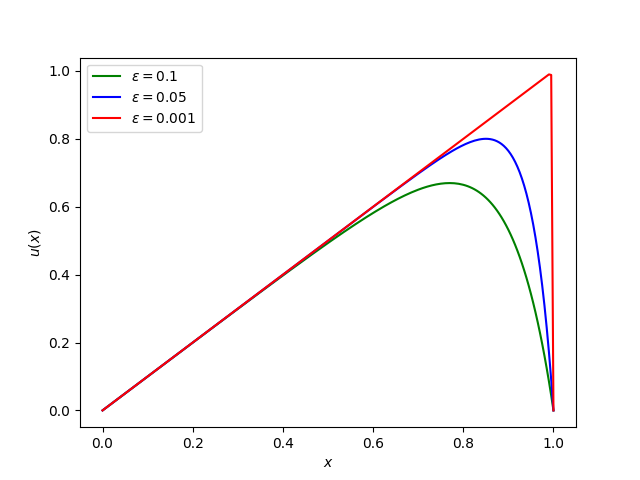}\label{fig:ad_example}}\hspace{5pt}
	\subfloat[]{\includegraphics[width=.4\columnwidth]{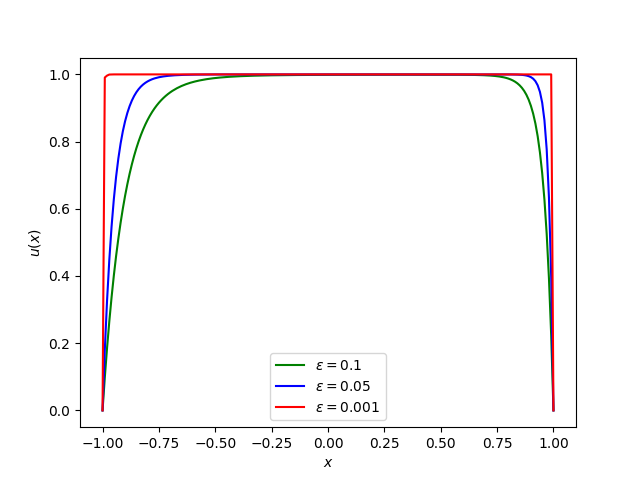}\label{fig:tp_example}}
	\caption{(a) Solutions to Eq. \eqref{equ:ode1} on $[0,1]$ with $b(x)=f(x)=1$ and $c(x)=0$. (b) Numerical solutions to Eq. \eqref{equ:ode_tp} with $b(x)=x(x+2)$ and $c(x)=f(x)=1$. There is a pronounced trend: as $\varepsilon$ decreases, solution $u$ sharpens near boundary layers.}
\end{figure}

Considering the one-dimensional singularly perturbed differential equations described as follows:
\begin{equation}
\left\{
\begin{array}{lr}
-\varepsilon u''+b(x)u'+c(x)u=f(x), & x\in(0,1), \\
u(0)=u(1)=0.
\end{array}
\right.
\label{equ:ode1}
\end{equation}
In Eq. \eqref{equ:ode1}, when the condition $b(x)>0$ holds on $[0,1]$, the solution typically displays an exponential boundary layer at $x=1$, while a similar layer emerges at $x=0$ when $b(x)<0$. For instance, considering the specific case where $b(x)=f(x)=1$ and $c(x)=0$, the solution is explicitly given by:
\begin{equation*}
u(x)=x-\frac{\exp\left(-\frac{1-x}{\varepsilon}\right)-\exp\left(-\frac{1}{\varepsilon}\right)}{1-\exp\left(-\frac{1}{\varepsilon}\right)},
\end{equation*}
this solution is visually represented in Fig. \ref{fig:ad_example}.

\begin{figure}[thbp]
    \centering
    \includegraphics[width=0.7\textwidth]{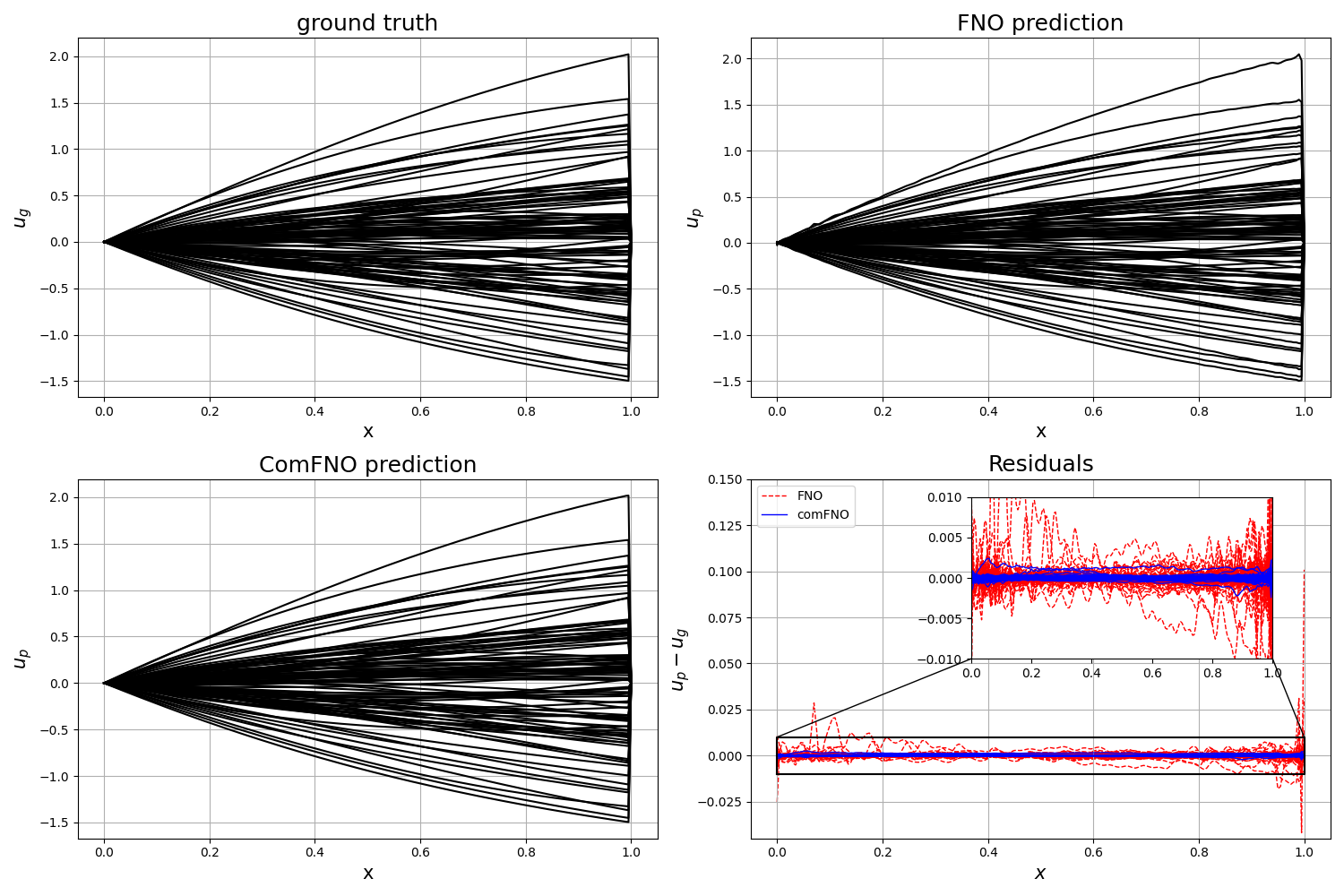}
    \caption{Performance of FNO and ComFNO on Eq. \eqref{eq:ordinary_1} with $\varepsilon=0.001$. In the upper-left figure, 100 curves depict the ground truth corresponding to 100 test $f$ samples. The remaining figures illustrate the predictions and residuals produced by FNO and ComFNO for these 100 $f$ samples.}
    \label{fig:residual_ode1}
\end{figure}

In cases where the coefficient of $u'$ has zeros, we refer to this as the turning points problem, and these zeros are termed turning points. Without loss of generality, we consider the following problem with a single turning point at $x=0$ over the interval $[-1,1]$:
\begin{equation}
\left\{
\begin{aligned}
&-\varepsilon u''+xb(x)u'+c(x)u=f(x),\quad x\in\left(-1,1\right),\\
&u(-1)=u(1)=0,
\end{aligned}\right.\label{equ:ode_tp}
\end{equation}
subject to the following hypotheses:
\begin{equation*}
b(x)\ne 0\ \text{on}\ [-1,1],\ c(x)\ge0,\ c(0)>0.\label{equ:tp_hypotheses}
\end{equation*}
It is crucial to emphasize that the solution $u(x)$ may demonstrate singular behavior at $x=0$ and the boundary points $x=-1$ and $x=1$. Specifically, when $b(-1)$ is negative, the solution $u(x)$ exhibits an exponential boundary layer at $x=-1$. Similarly, if $b(1)$ is positive, $u(x)$ displays an exponential boundary layer at $x=1$. Moreover, the behavior of the solution $u(x)$ is influenced by the parameter $\lambda=c(0)/b(0)$. In the case of $\lambda<0$, an inner layer emerges at $x=0$, whereas for $\lambda>0$, $u(x)$ remains smooth near $x=0$. As an example, the numerical solution to Eq. \eqref{equ:ode_tp} with $b(x)=x(x+2)$, $c(x)=f(x)=1$, using the upwind scheme on the Shishkin mesh, is shown in Fig. \ref{fig:tp_example}.

\begin{figure}
    \centering
    \includegraphics[width=0.7\textwidth]{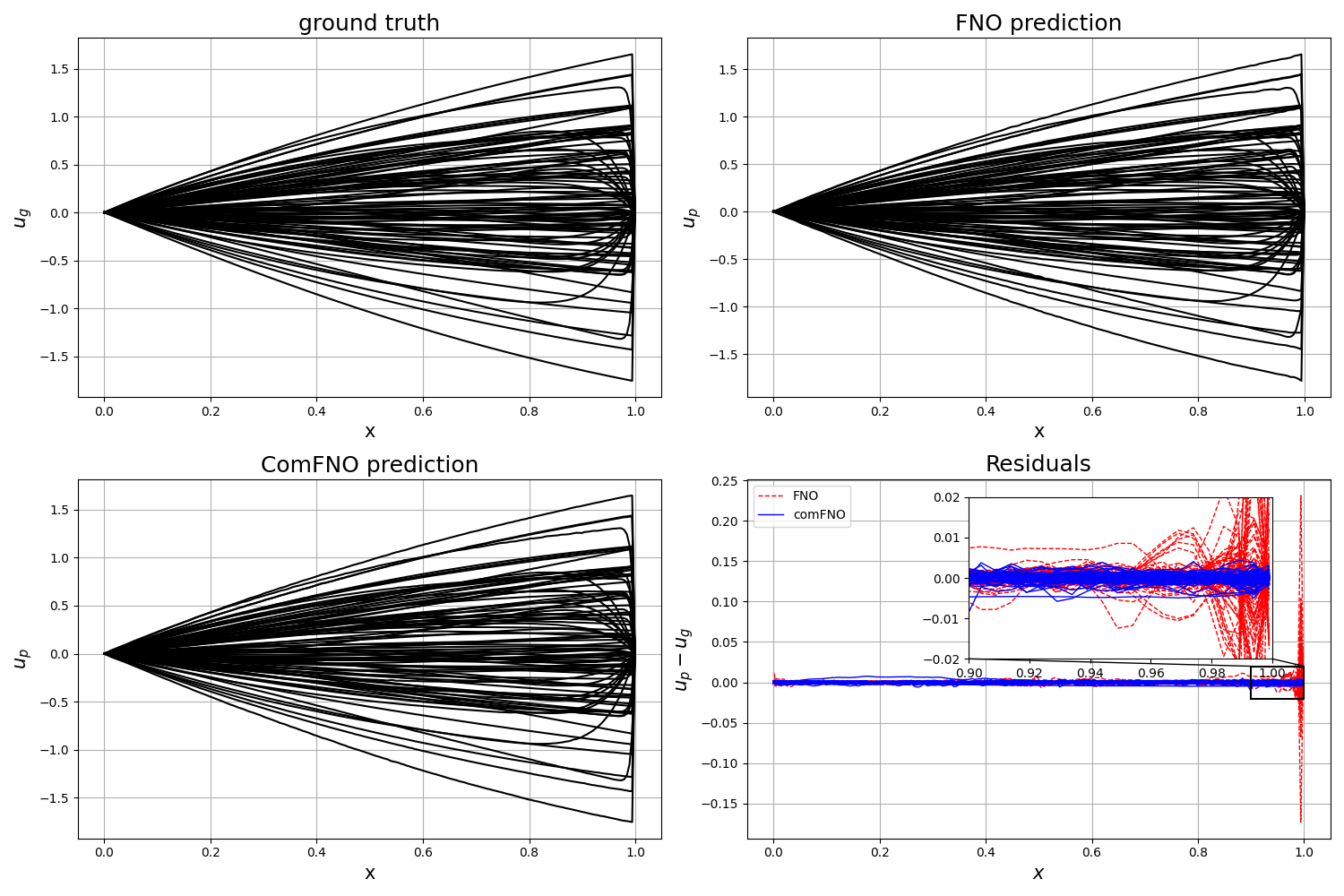}
    \caption{(Case of multiple data distributions) Performance of both FNO and ComFNO on Eq.\eqref{eq:ordinary_1}. Both models are evaluated across 100 $f$ samples, each featuring distinct $\varepsilon$. The resulting prediction and residual curves are depicted accordingly.}
    \label{fig:residual_mul_eps}
\end{figure}

Subsequently, we employ ComFNO and FNO on the aforementioned equations. We commence with the subsequent problem:
\begin{equation}
\left\{
    \begin{aligned}
    &-\varepsilon u''+(x+1)u'=f,\quad x\in(0,1),\\
    &u(0)=u(1)=0.
    \end{aligned}\right.
    \label{eq:ordinary_1}
\end{equation}
The trained FNO model and ComFNO model effectively serve as solvers for partial differential equations. Assessing both trained models on 100 randomly selected $f$ samples with a resolution of 201, the results encompassing ground truth, FNO predictions, ComFNO predictions, as well as the respective residuals, are depicted in Fig. \ref{fig:residual_ode1}.

\begin{figure}
    \centering
    \includegraphics[width=0.7\textwidth]{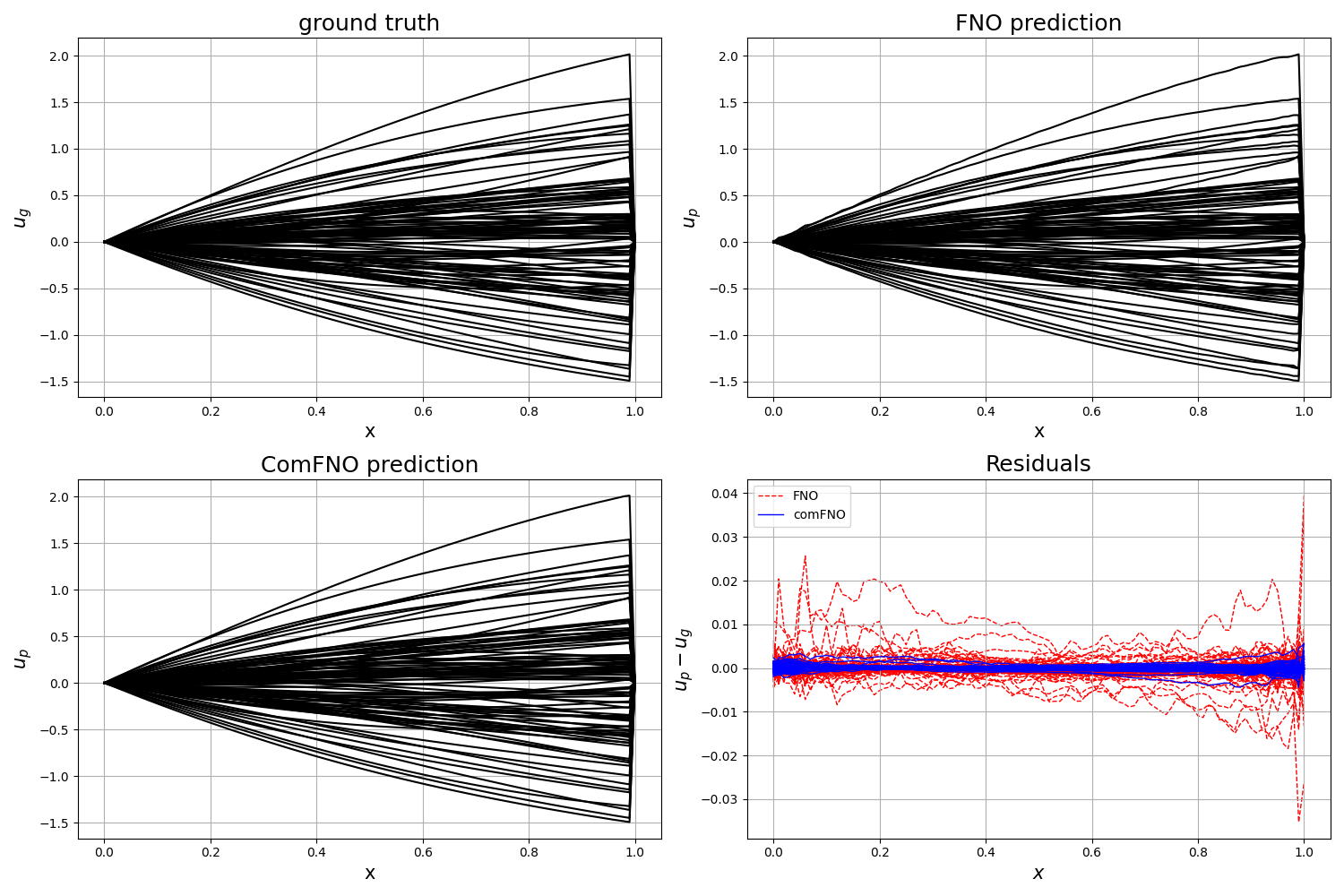}
    \caption{(Case of few-shot) Performance of FNO and ComFNO on Eq. \eqref{eq:ordinary_1} with $\varepsilon=0.001$. In the upper-left figure, 100 curves depict the ground truth corresponding to 100 test $f$ samples. The remaining figures illustrate the predictions and residuals produced by FNO and ComFNO for these 100 $f$ samples.}
    \label{fig:residual_fewshot}
\end{figure}

Continuing our examination of Eq. \eqref{eq:ordinary_1}, we now explore the scenario involving multiple data distributions. The outcomes of this investigation are presented in Fig. \ref{fig:residual_mul_eps}. Additionally, we delve into the few-shot case, with the corresponding results showcased in Fig. \ref{fig:residual_fewshot}.

\begin{figure}
    \centering
    \includegraphics[width=0.7\textwidth]{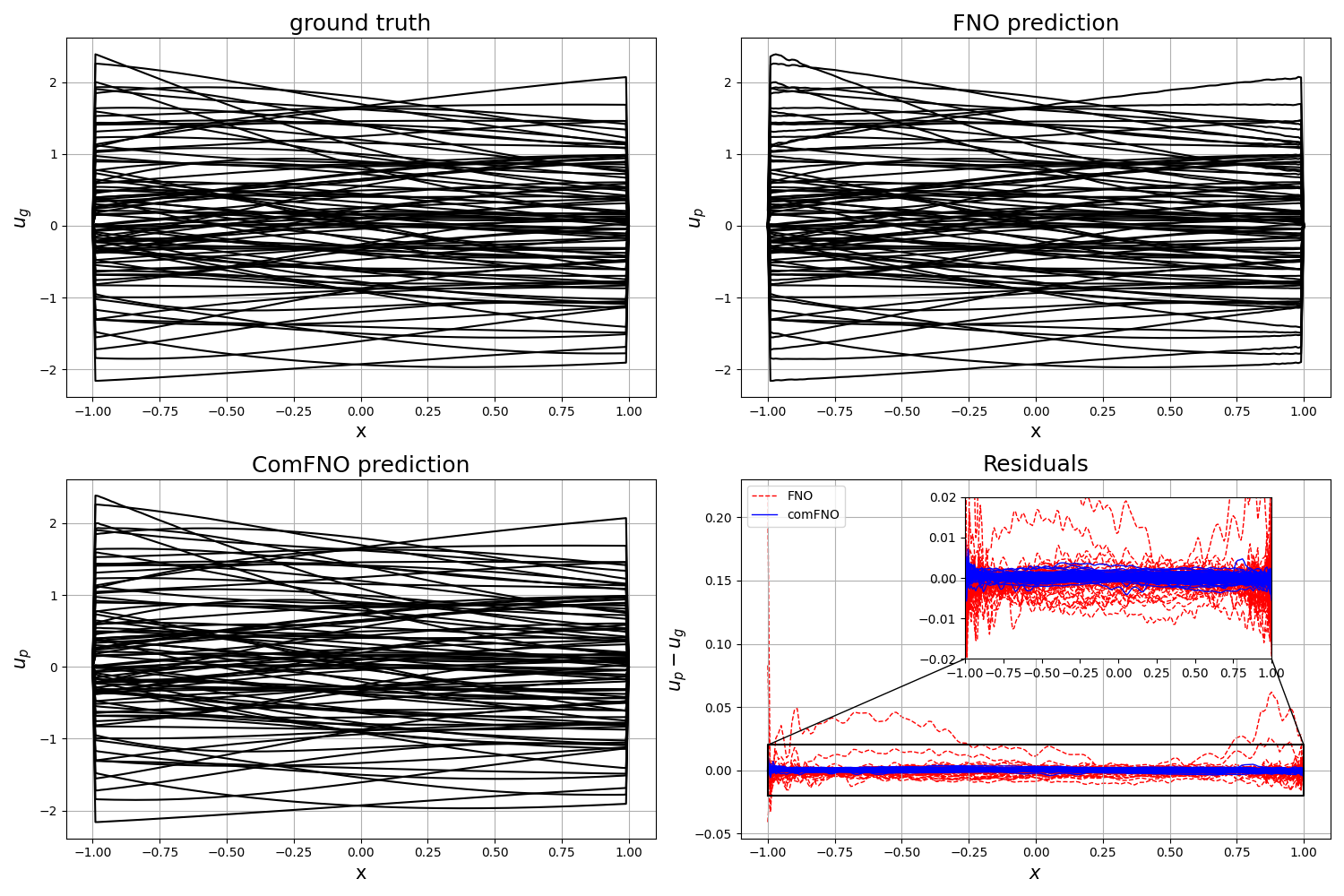}
    \caption{Performance of FNO and ComFNO on Eq. \eqref{equ:tp} with $\varepsilon=0.001$. In the upper-left figure, 100 curves depict the ground truth corresponding to 100 test $f$ samples. The remaining figures illustrate the predictions and residuals produced by FNO and ComFNO for these 100 $f$ samples.}
    \label{fig:residual_tp}
\end{figure}

Subsequently, we address the subsequent turning point problem:
\begin{equation}
\left\{
    \begin{aligned}
    &-\varepsilon u''+x(x+2)u'+u=f,\quad x\in(-1,1),\\
    &u(-1)=u(1)=0.
    \end{aligned}\right.
    \label{equ:tp}
\end{equation}
The solutions to this problem exhibit two boundary layers at both $x=-1$ and $x=1$. We employ ComFNO and FNO to solve this problem, presenting the ground truth, predictions, and residuals of both models for the test dataset in Figure \ref{fig:residual_tp}.

\subsection{One-dimensional Time-dependent Problems}
\begin{figure}[htbp]
    \centering
    \includegraphics[width=0.9\textwidth]{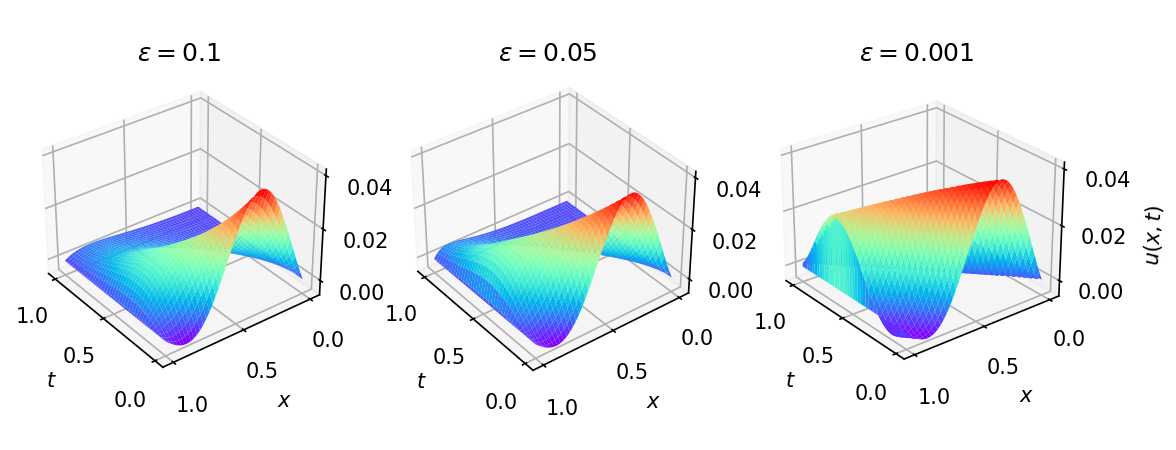}
    \caption{The numerical solution to Eq. \eqref{eq:ib} associated with a random $f$. As $\varepsilon$ decreases, the solution demonstrates intensified rapid transitions in the proximity of $x=1$.}
    \label{fig:ib_example}
\end{figure}

\begin{figure}[htbp]
    \centering
    \includegraphics[width=0.8\textwidth]{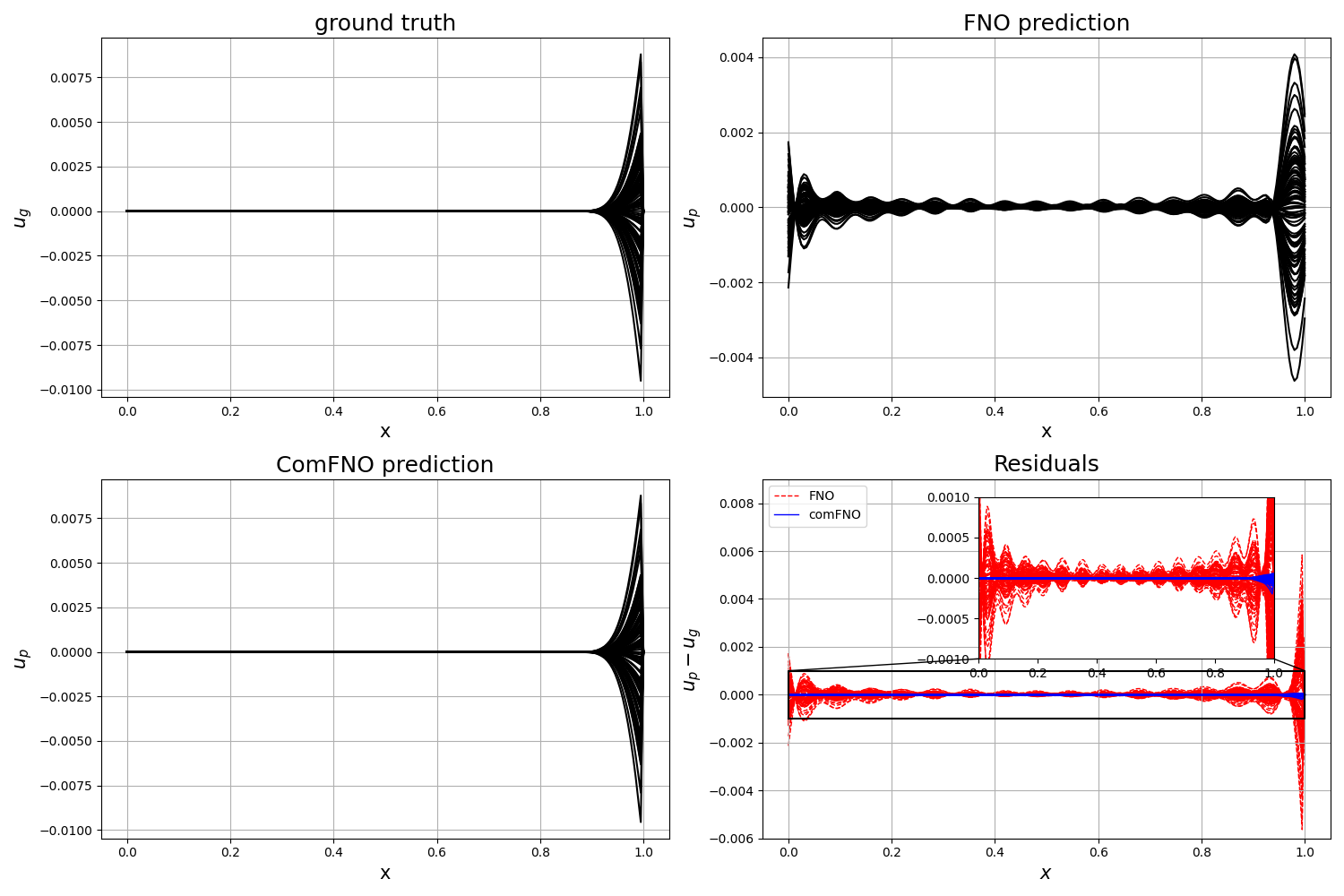}
    \caption{Performance of FNO and ComFNO on Eq. \eqref{eq:ib} with $\varepsilon=0.001$. In the upper-left figure, 100 curves depict the ground truth corresponding to 100 test $f$ samples. The remaining figures illustrate the predictions and residuals produced by FNO and ComFNO for these 100 $f$ samples.}
    \label{fig:ib_residual}
\end{figure}

In the context of parabolic partial differential equations in the space-time domain $Q=(0,1)\times(0,T]$, the initial-boundary value problem is described by the following equation:
\begin{equation}
\left\{
    \begin{aligned}
    &u_t-\varepsilon u_{xx}+b(x,t)u_x+d(x,t)u=f(x,t),\ &(x,t)\in Q,\\
    &u(x,0)=s(x),\quad&0\le x\le 1,\\
    &u(0,t)=q_0(t),\quad&0< t\le T,\\
    &u(1,t)=q_1(t),\quad&0< t\le T.
    \end{aligned}\right.\label{pde:ib}
\end{equation}
In cases where $b > 0$, the solution $u$ displays smooth behavior across most of domain $Q$. However, near the boundary $x = 1$ of $Q$, the solution typically manifests a boundary layer. Remarkably, the mapping $x\mapsto 1-x$ transforms the case of $b < 0$ to that of $b > 0$. For fixed $t > 0$, the behavior of this boundary layer concerning $x$ mirrors that depicted in Eq. \eqref{equ:ode1}. As an illustrative example, consider the following problem:
\begin{equation}
\left\{\begin{array}{lr}
    u_t-\varepsilon u_{xx}+u_x+xu=0, &  (x,t)\in (0,1)\times(0,1],\\
    u(x,0)=f(x), & x\in [0,1],\\
    u(0,t)=u(1,t)=0,& t\in [0,1].
\end{array}\right.
\label{eq:ib}
\end{equation}
The solution presents a boundary layer near $x=1$, as evident from the numerical solutions obtained using the Crank-Nicolson scheme on the Shishkin mesh, depicted in Fig. \ref{fig:ib_example}. We proceed to employ ComFNO and FNO on Eq. \eqref{eq:ib}, presenting the ground truth, predictions, and residuals for both models in Fig. \ref{fig:ib_residual}.


\subsection{Two-dimensional Problems}
For a boundary value problem of an elliptic partial differential equation in the spatial domain $\Omega=(0,1)\times(0,1)$, given by
\begin{equation}
\left\{
    \begin{aligned}
    &-\varepsilon\Delta u+\textbf{b}(x,y)\cdot\nabla u+c(x,y)u =f(x,y),\ &\text{in}\ \Omega,\\
    &u(x,y)=0,\  &\text{on}\ \partial\Omega.
    \end{aligned}\right.\label{pde:2d}
\end{equation}
Under the assumption $\textbf{b}=(b_1,b_2)>0$ (specifically, with $b_1>0$ and $b_2>0$), the solution to this problem features boundary layers along $x=1$ and $y=1$, with their intersection occurring at the corner point $(1, 1)$, thereby leading to the emergence of a corner layer. 

\begin{figure}[htbp]
    \centering
    \includegraphics[width=0.7\textwidth]{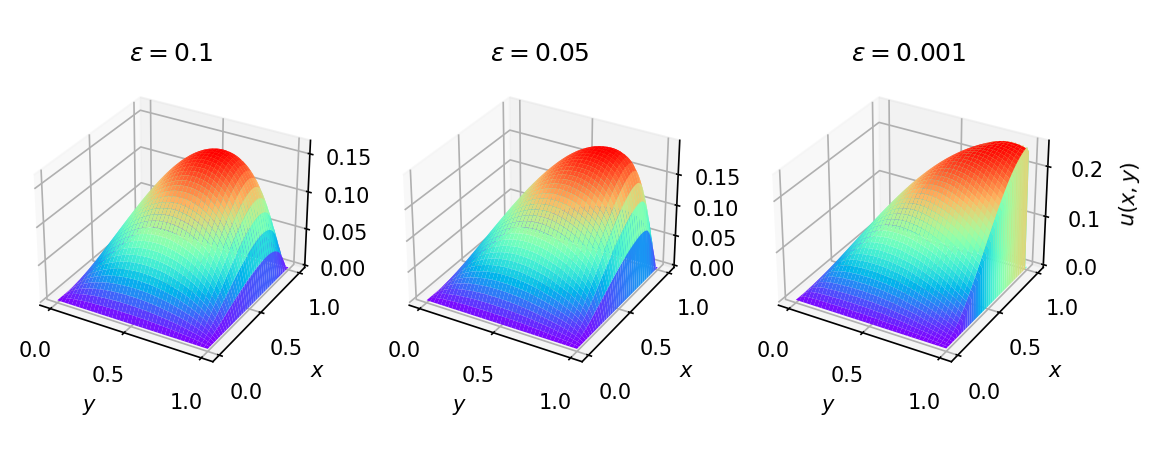}
    \caption{The numerical solution to Eq. \eqref{eq:partial_2} with a random $f$. As $\varepsilon$ decreases, the solution demonstrates intensified rapid transitions in the proximity of $x=1$ and $y=1$.}
    \label{fig:2d_example}
\end{figure}

\begin{figure}[htbp]
	\centering
	\subfloat[]{\includegraphics[width=.9\columnwidth]{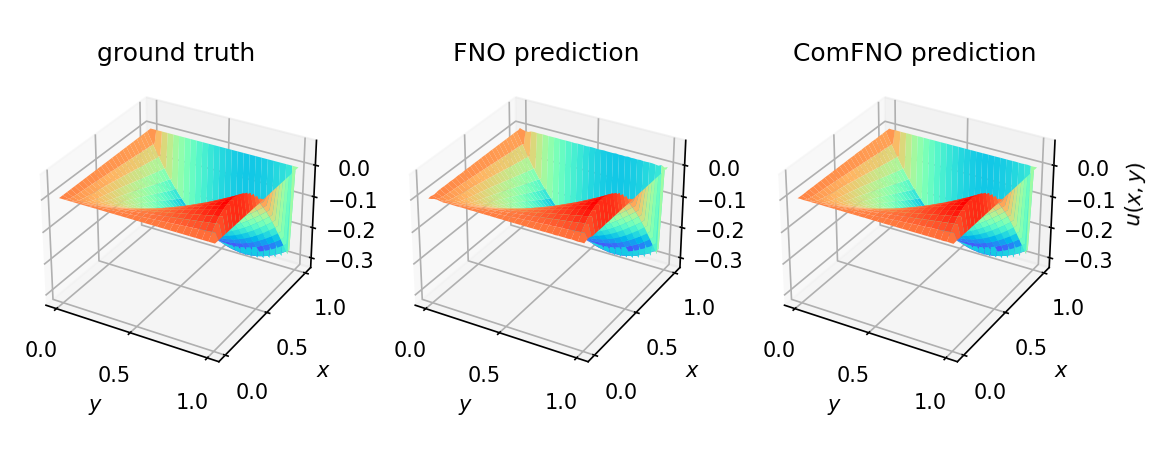}}\\
	\subfloat[]{\includegraphics[width=.7\columnwidth]{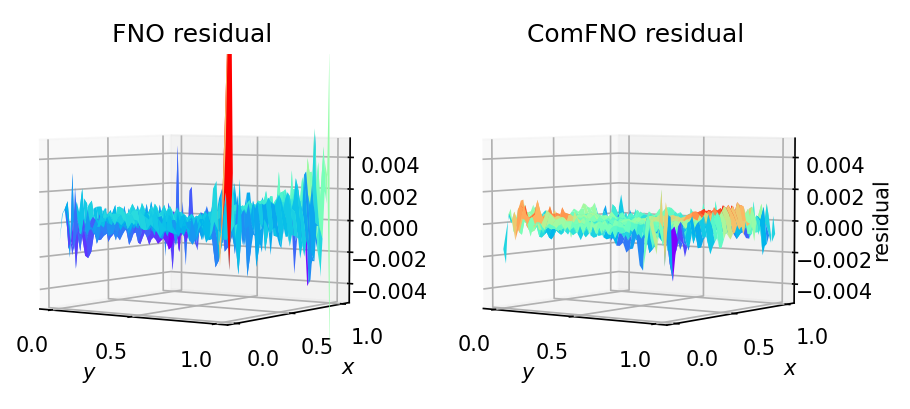}}\hspace{5pt}
	\caption{Depicting the performance of the trained FNO and trained ComFNO models on Equation \eqref{eq:partial_2} with $\varepsilon=0.001$, considering a random $f$ sample.}
 \label{fig:2d_residual}
\end{figure}

As an illustrative instance, consider the subsequent problem:
\begin{equation}
\left\{
\begin{array}{lr}
    -\varepsilon\Delta u+u_x+u_y+u=f(x), &  (x,y)\in(0,1)\times(0,1),\\
    u(0,y)=u(1,y)=0, &y\in [0,1],\\
    u(x,0)=u(x,1)=0, &x\in [0,1].\\
\end{array}\right.
\label{eq:partial_2}
\end{equation}
For a visual representation of layers, one can refer to Fig. \ref{fig:2d_example}. To address Eq. \eqref{eq:partial_2}, we employ both ComFNO and FNO. The ensuing outcomes, encompassing the ground truth, predictions, and residuals on a random $f$ sample for both models, are presented in Fig. \ref{fig:2d_residual}.

\end{document}